\documentclass[10pt,twocolumn,letterpaper]{article}

\usepackage{iccv}              

\definecolor{iccvblue}{rgb}{0.21,0.49,0.74}
\usepackage[pagebackref,breaklinks,colorlinks,allcolors=iccvblue]{hyperref}
\usepackage{amsmath,bm}
\usepackage{multirow}
\usepackage{url}            
\usepackage{booktabs}       
\usepackage{amsfonts}       
\usepackage{nicefrac}       
\usepackage{microtype}      
\usepackage{xcolor}         
\usepackage{graphicx}
\usepackage{enumitem}
\usepackage{color}
\usepackage{soul}
\usepackage{makecell}

\def\paperID{7586} 
\def\confName{ICCV}
\def\confYear{2025}

\def\name{PBR3DGen}
\title{PBR3DGen: A VLM-guided Mesh Generation with High-quality PBR Texture  }
\author{
Xiaokang Wei\textsuperscript{1,4}\footnotemark[1] \quad
Bowen Zhang\textsuperscript{2,4}\footnotemark[1] \quad
Xianghui Yang\textsuperscript{4} \quad
Yuxuan Wang\textsuperscript{3,4}\\
Chunchao Guo\textsuperscript{4}\quad
Xi Zhao\textsuperscript{2} \quad
Yan Luximon\textsuperscript{1}\thanks{Corresponding Author} \\
\textsuperscript{1}The Hong Kong Polytechnic University \quad 
\textsuperscript{2}Xi'an Jiaotong University \quad \\
\textsuperscript{3}Nanyang Technological University \quad 
\textsuperscript{4}Tencent Hunyuan
}

\begin{document}

\twocolumn[{%
\renewcommand\twocolumn[1][]{#1}%
\maketitle
\vspace{-0.2cm}
\includegraphics[width=\linewidth]{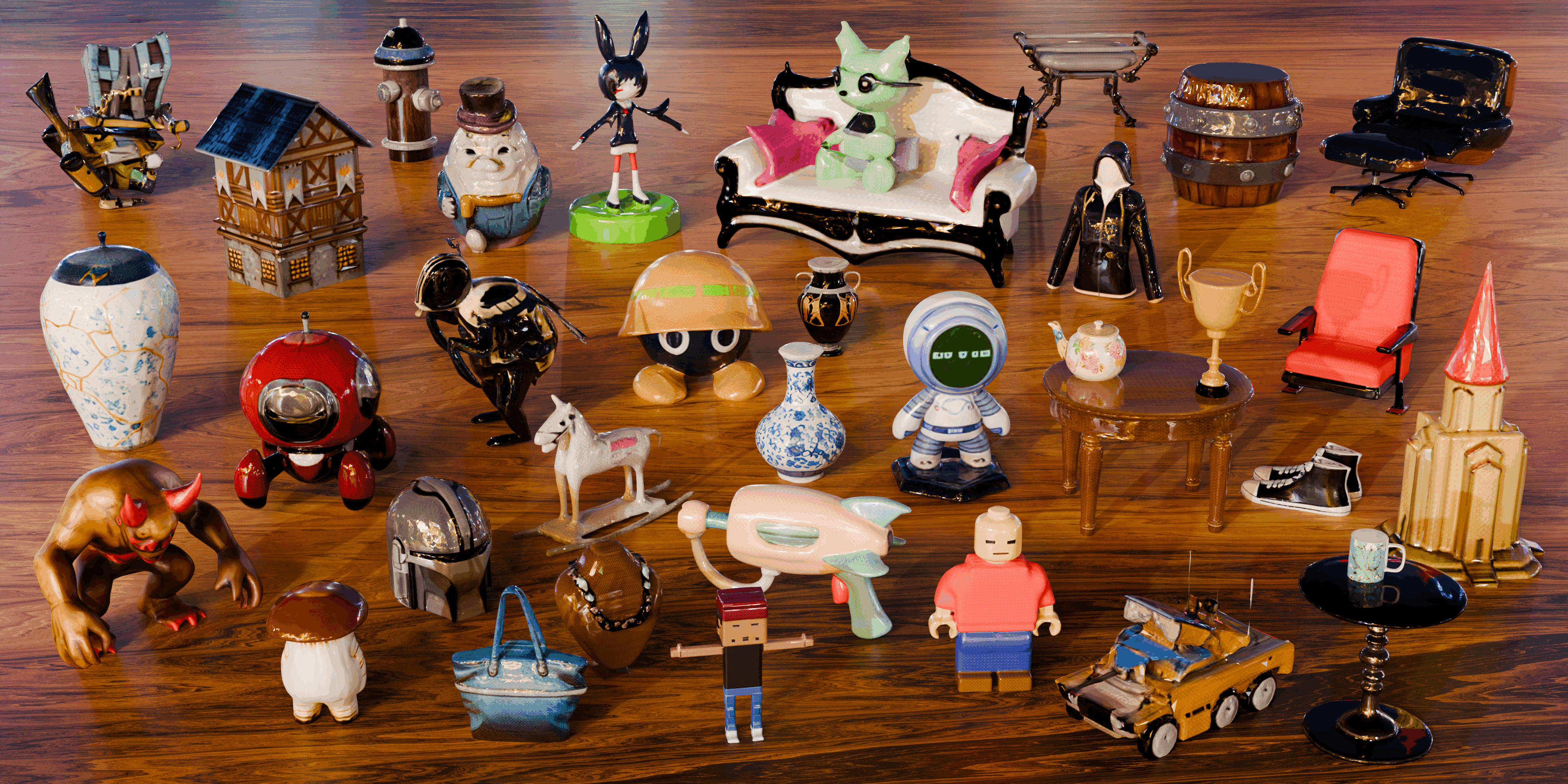}
\captionof{figure}{We present \textbf{PBR3DGen}, a novel two-stage 3D assets generation framework with high-quality physically-based rendering materials. All objects in the scene are generated from PBR3DGen. \vspace{1em}}
\label{fig:teaser}
}]


\begin{abstract}
Generating high-quality physically based rendering (PBR) materials is important to achieve realistic rendering in the downstream tasks, yet it remains challenging due to the intertwined effects of materials and lighting.
While existing methods have made breakthroughs by incorporating material decomposition in the 3D generation pipeline, they tend to bake highlights into albedo and ignore spatially varying properties of metallicity and roughness. In this work, we present {\bf~\name{}}, a two-stage mesh generation method with high-quality PBR materials that integrates
the novel multi-view PBR material estimation model and a 3D PBR mesh reconstruction model. Specifically,~\name{} leverages vision language models (VLM) to guide multi-view diffusion, precisely capturing the spatial distribution and inherent attributes of reflective-metalness material.
Additionally, we incorporate view-dependent illumination-aware conditions as pixel-aware priors to enhance spatially varying material properties. Furthermore, our reconstruction model reconstructs high-quality mesh with PBR materials. Experimental results demonstrate that \name{} significantly outperforms existing methods, achieving new state-of-the-art results for PBR estimation and mesh generation. More results and visualization can be found on our project page: 
\href{https://pbr3dgen1218.github.io/}{https://pbr3dgen1218.github.io/}.
\end{abstract}

\section{Introduction}

Generating high-quality 3D mesh with physically based rendering (PBR) materials from images or text prompts has broad applications such as 3D graphics pipelines, movie production, gaming and AR/VR. 3D generation models have witnessed remarkable progress which can be attributed to the scalability of 3d generative models~\cite{Tang2023mvdiffusion,long2024wonder3d,shi2023zero123++,xu2024instantmesh}, and utilization of the large-scale training datasets~\cite{deitke2023objaverse}. However, most existing 3D mesh and texture generation models often lack PBR material properties so they lose the view-dependent photorealistic effect. Furthermore, the textures, which incorporate pre-baked shadows and lighting, restrict their applicability in downstream tasks.

PBR estimation from images is a recent development in the field of 3D assets with PBR generation, such as Clay~\cite{zhang2024clay} has demonstrated impressive capabilities in 3D mesh with PBR materials generation, but they rely on expensive geometric prior to ensure cross-view PBR consistency. 3DTopia-XL~\cite{chen20243dtopia} encodes detailed shape, Albedo, and material field into a Diffusion Transformer (DiT) framework. To achieve better performance, these methods have attempted to incorporate BSDF and illumination models into 3D generation model. More recently, Meta3DAssetGen~\cite{siddiqui2024meta} exploit to integrate the differentiable BRDF optimization model into forward transformer-based architecture. SF3D~\cite{boss2024sf3d} incorporates explicit lighting and a differentiable shading model for decomposing light from UV texture. 

In the field of 3D generation, PBR material representation has recently emerged as a key advancement, significantly improving the rendering quality of generated objects. For example, Clay~\cite{zhang2024clay} has demonstrated impressive capabilities to generate PBR materials for 3D meshes, yet it requires geometric input to maintain cross-view consistency and concurrently suffers from the deterioration of detailed textures
3DTopia-XL~\cite{chen20243dtopia} develops a primitive presentation to encode detailed shape, Albedo, and material fields into a Diffusion Transformer (DiT) framework. On the other hand, recent methods have attempted to incorporate BSDF material functions and illumination models into 3D generation models. Meta3DAssetGen~\cite{siddiqui2024meta} exploits the integration of the differentiable BRDF optimization model into a large reconstruction model (LRM) with a forward transformer-based architecture. SF3D~\cite{boss2024sf3d} incorporates explicit lighting and a differentiable shading model for decomposing light from UV textures.

However, by simplifying material models with a single illumination for different objects and ignoring the spatially varying properties of metalness and roughness, these methods face three significant limitations that hinder their widespread adoption. Firstly, these methods are more likely to bake high light into the Albedo map due to the existing ambiguity between illumination and Albedo, especially for reflective objects.
Secondly, roughness and metallicity are difficult to observe directly from RGB images. We also noticed that the BRDF distribution in current synthetic 3D datasets, such as Objaverse~\cite{deitke2023objaverse}, exhibits a strong long-tail effect. This causes models to overfit to frequent values while ignoring rare ones. As a result, the quality of PBR material generation is compromised, and spatially-varying attributes are poorly represented.
Additionally, Meta3DAssetGen~\cite{siddiqui2024meta} employs LRM to predict PBR materials, which is trained from scratch. Given the scarcity of high-quality 3D PBR data, the generalization capability of the LRM may not be as strong as that of diffusion models for predicting PBR. Meanwhile, the performance of sparse-view reconstruction models can decline if the quality of multiple views is poor, as these models generally depend exclusively on view-aware RGB images to predict 3D representations.

In response to these challenges, we propose an efficient approach for high-quality 3D mesh with PBR materials generation from a single image or text prompt that disentangles the highlight and reconstructs spatially-varying metallic and roughness to enable relighting (see Fig.\ref{fig:teaser}). Our method introduces PBR3DGen, a two-stage 3D mesh generation method with high-fidelity PBR materials that integrates the novel PBR multiview diffusion models and PBR-based sparse-view reconstruction models to achieve high-quality 3D mesh with PBR materials generation. In the first stage, we leverage vision language models (VLM) like GPT-4V and view-dependent illumination-aware conditions to guide our multi-view PBR estimation model. Specifically, we employ VLM to precisely capture the spatial distribution and inherent attributes of reflective-metallic materials. This detailed information is then seamlessly integrated into a PBR multi-view diffusion model. This integration plays a pivotal role in drastically mitigating the ambiguity often encountered between specular highlights and Albedo within the rendered imagery. Furthermore, it addresses the issue of prediction inaccuracies in metallic and roughness properties, which were previously exacerbated by the severe long-tail distribution of training data. Consequently, our approach significantly enhances the consistency of part-aware material representation. In addition, we inject view-dependent illumination-aware conditions to enhance the spatially varying material properties.
In the second stage, unlike most sparse-view reconstruction models that reconstruct 3D assets using only rgb color, we propose our PBR-based large reconstruction model, which employs a dual-head VAE encoder to separately encode the Albedo and Metallic-Roughness maps. Subsequently, we reconstruct the 3D mesh and PBR materials from the input PBR multi-view images.

In summary, our key contributions to the field of PBR 3D reconstruction are as follows:
\begin{itemize}[leftmargin=*]
\setlength{\itemsep}{1pt}
\setlength{\parsep}{1pt}
\setlength{\parskip}{1pt}
    \item We propose PBR3DGen, a novel two-stage generation framework for 3D assets with high-quality PBR materials from image or text inputs; 
    \item We explore leveraging vision-language models to guide PBR estimation within a multi-view diffusion model, enabling more accurate estimation of PBR materials and significantly reducing ambiguities between Albedo and illumination, especially specular highlights.
    \item We introduce a view-dependent illumination-aware condition as a local pixel-wise prior, resulting in a more accurate capture of spatially varying reflectance effects.
    \item Our method exhibits superior 3D mesh and PBR material generation quality compared to current methods.
\end{itemize}


\section{Related Works}

\subsection{Multi-view Diffusion}
Cross-view consistency is crucial in a reconstruction-based 3D generation. MVDiffusion\cite{Tang2023mvdiffusion} first generates consistent multi-view images from text prompts, given pixel-to-pixel correspondences. SyncDreamer\cite{liu2023syncdreamer}, MVDream\cite{shi2023mvdream}, and Wonder3D\cite{long2024wonder3d} leverage attention mechanisms to facilitate information transfer between multi-view images, enhancing multi-view consistency. Zero123++\cite{shi2023zero123++} stitches multi-view images together while denoising them simultaneously, improving geometric consistency and texture quality. Era3D\cite{li2024era3d} introduces row-wise multi-view attention, reducing the computational overhead of multi-view generation. Although the quality of current multi-view image generation has made significant progress, multi-view images embed lighting and lack physical properties. In contrast to these previous methods, our approach not only maintains multi-view consistency but also provides PBR estimation. This enables our reconstruction results to support relighting and physically-based rendering.

\subsection{Multi-view 3D Reconstruction}
3D reconstruction has been a well-researched area in the field of computer vision for a long time. Although traditional methods such as Structure from Motion (SfM)\cite{agarwal2011building, pollefeys2004visual, schonberger2016structure} and Multi-View Stereo (MVS)\cite{furukawa2009accurate, pollefeys2008detailed, schonberger2016pixelwise} can perform camera calibration and 3D reconstruction, they lack robustness when dealing with inconsistent multi-view images. Recently, deep learning-based 3D reconstruction methods have become mainstream.
LRM\cite{hong2023lrm} first proposed utilizing a transformer backbone to simultaneously reconstruct geometry and texture through a single forward pass. LRM can learn how to reconstruct 3D geometry and texture from a single image using large-scale 3D datasets. Instant3D\cite{li2023instant} increases the number of input views, further enhancing the geometric detail and texture quality of the reconstruction. Subsequent works\cite{zhang2024geolrm, wei2024meshlrm, zhang2025gs, xu2024instantmesh, li2022meshformer} have made further improvements in reconstruction quality and computational efficiency.
Compared to existing large reconstruction models, we not only reconstruct the 3D geometry, but we also reconstruct physics-based material properties, making the 3D assets generated by our method more realistic.

\subsection{Diffusion-based PBR Material Generation}
Material generation and estimation from RGB images is inherently difficult because of its under-constrained nature, including the ambiguity between illumination and materials.
Diffusion models reveal the impressive capability of learning
the distribution of the target domain, which has become prominent in texture content generation. Recent studies such as TEXTure~\cite{richardson2023texture}, Text2Tex~\cite{chen2023text2tex}, TexFusion~\cite{cao2023texfusion} extend the diffusion model to texture synthesize from multi-view images. SyncMVD~\cite{liu2023text} improves the consistency of multi-view texture by sharing the denoised content among different views in each denoising step to ensure texture consistency and avoid seams and fragmentation.  However, these methods tend to generate RGB textures with highlights and shadows due to suffering from disentangling the materials and illumination properties. Recently, Paint3D~\cite{zeng2024paint3d} proposed a coarse-to-fine strategy to delight the generated texture in UV space, but they still lack Physically-based materials when dealing with inconsistent multi-view images. 
Fantasia3D~\cite{chen2023fantasia3d} can generate more realistic textures by incorporating physics-based materials. FlashTex~\cite{deng2024flashtex} introduces the lighting condition in ControlNet and optimizes texture based on Score Distillation Sampling loss, which can disentangle lighting from surface material/reflectance. However, these optimization-based methods require extensive training time. RGBX~\cite{zeng2024rgb} utilizes diffusion models to estimate the Physically-based intrinsics of RGB images and demonstrates a significant improvement on the generalization. Additionally,  IntrinsicAnything~\cite{chen2024intrinsicanything} only utilizes diffusion models to estimate albedo and specular shading from a single RGB image, which lacks roughness and metallic properties in 2D diffusion prior stage. Meanwhile, these methods still suffer from disentangling the spatially varying materials from highlight and reflective object images.

\section{Methods}

\subsection{Overview}
\begin{figure*}[t]
\centering
  \includegraphics[width=0.96\linewidth]{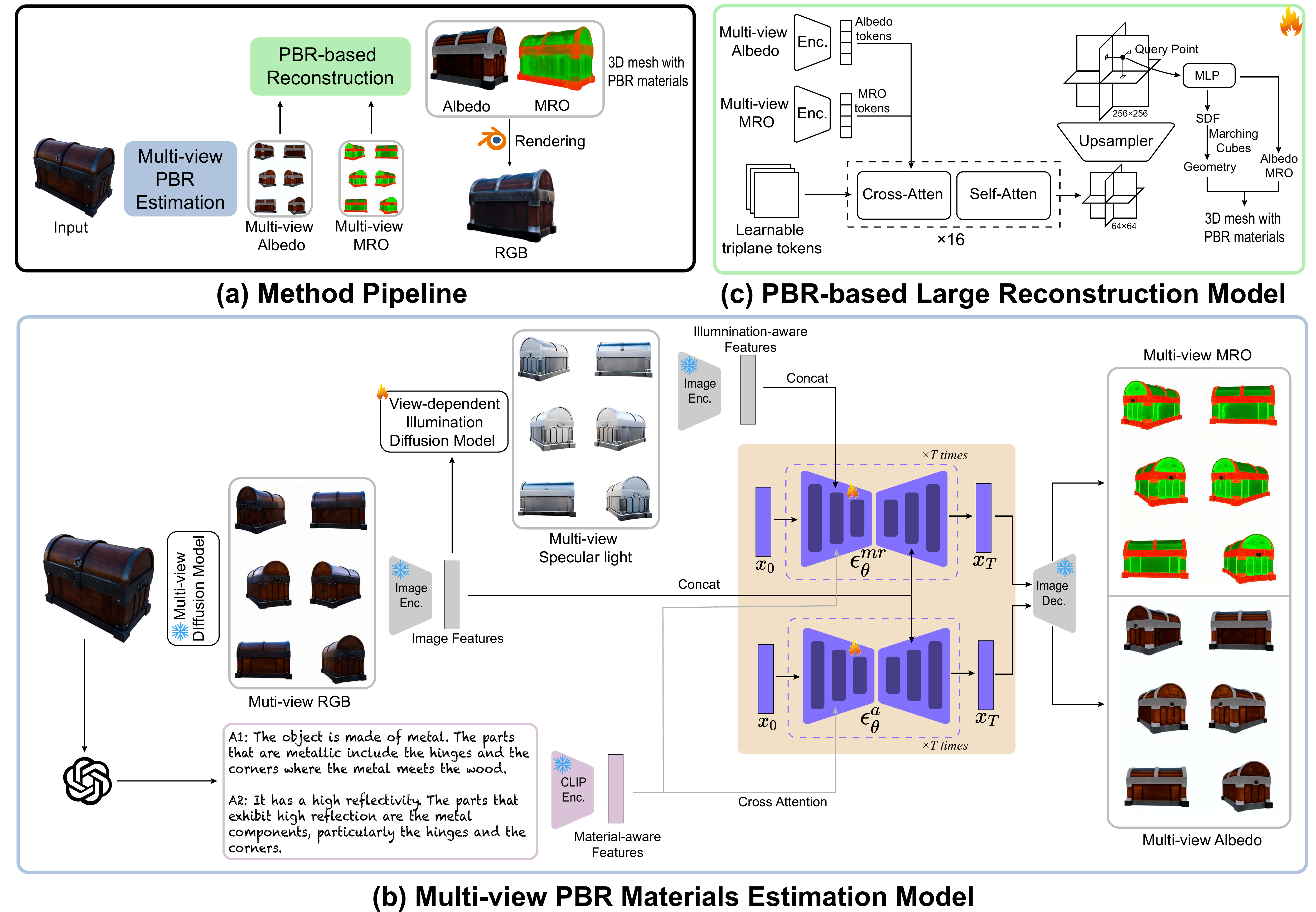}
  \vspace{-0.2cm}
\caption{Method overview. Our method consists of two stages: Multi-view PBR materials estimation and 3D mesh with PBR materials reconstruction. Given an RGB image as input, we first generate multi-view Albedo images and multi-view MRO images using Multi-view PBR estimation model, and then reconstruct 3D assets with Dual-head PBR-based large reconstruction model. }
\label{fig:overview}
\vspace{-0.4cm}
\end{figure*}

In this section, we introduce our high-quality 3D mesh with PBR materials generation pipeline, illustrated in Fig.\ref{fig:overview}. Firstly, we establish a novel multi-view PBR texture diffusion model from a single image or text prompt, which unleashes Vision Language model (VLM) and illumination-aware conditions to guide PBR material generation. Secondly, we further design dual-head reconstruction model by extending a sparse-view large reconstruction model to handle multi-view PBR material inputs, significantly boosting the 3D mesh and texture reconstruction quality.

\subsection{Multi-view PBR Estimation}
Without prior geometric information, generating multi-view PBR materials directly from a single RGB image using a diffusion model is a challenging problem. Due to the inherent ambiguities between material properties and lighting conditions, and the observation that the local distribution of material characteristics does not always align with geometric consistency. Multi-view PBR estimation from an RGB image cannot achieve both high-quality PBR materials and geometrically consistent multi-view images simultaneously. Therefore, We first resolve multi-view consistency by applying an off-the-shelf Multi-view Diffusion Model, such as Zero123++\cite{shi2023zero123++}. Meanwhile, we propose a PBR generation module from the multi-view RGB image input by introducing VLM guidance and view-dependent illumination-aware condition, which can significantly handle the ambiguity between PBR and illumination well.

\noindent\textbf{Multi-view PBR Material Estimation with SD.} Since PBR material estimation can be seen as domain transfer from an RGB image, The diffusion model can be utilized to transfer from the RGB domain distribution to the PBR domain distribution. And we observed that Roughness and Metallic often interact to produce specular reflection effects on objects. In our methods, we combined two maps into a single MRO (Metallic/Roughness/Zero channel) for generation. Therefore, we can model the conditional distribution of corresponding Albedo and Metallic-Roughness by utilizing the RGB image as the conditioning signal, as in IntrinsicAnything~\cite{chen2024intrinsicanything} and RGBX~\cite{zeng2024rgb}. Specifically, we first use the pre-trained VAE image encoder $\mathcal{E}$ to extract the conditional signal feature from input grid image $\bm{I}$. Then, the diffusion process adds noise to the encoded latent $z = \mathcal{E}(x)$  producing a noisy latent $z_t$ where the noise level increases over timesteps $\bm{x} \in T $. We learn network $\epsilon_{\theta}$ that predicts the noise added to the noisy latent $z_t$ given image conditioning $\mathcal{E}(\bm{I})$.  We minimize the following loss:
\begin{equation}
\small
    L = \mathbb{E}_{\mathcal{E}(x),\mathcal{E}(I),\epsilon, t}\left[\left|\epsilon-\epsilon_{\theta}\left(z_{t}, t,\mathcal{E}(\bm{I})\right)\right|_{2}^{2}\right]\label{eq:1}
\end{equation}
\noindent where $z_{t}$ is the noisy latent feature of the input $z$ with t uniformly sampled from $\{1, \ldots, T\}$, and estimating $\epsilon$ from a Gaussian distribution, denotes $\epsilon\sim\mathcal{N}(0,1)$. Here, we separately optimize double U-Net Network $\epsilon_{\theta}^{a}$ and $\epsilon_{\theta}^{mr}$ corresponding Albedo estimation and MRO estimation.

\subsubsection{VLM-guided PBR Material generation}\label{sec:VLM-guided}
We observed that the Roughness and Metallic distributions for objects tend to be strongly part-aware consistent, and there are significant ambiguities between Albedo and specular light during the Albedo generation process, especially for specular and metallic objects. This is due to the high variance in the diffusion inference procedure.

As argued above, the PBR material generation model requires a strong prior to alleviate these challenging ambiguities. Inspired by \cite{fang2024make}, VLM can recognize object materials and types due to its extensive prior knowledge of objects. We design a hierarchical VLM-guided material policy to obtain reflective-metallic material attributes through unleashing GPT-4V with a strong material knowledge, which helps us capture the reflective-metallic properties for the input RGB image. Specifically, we first gather global information about metallic and reflective properties, and then we further investigate which parts exhibit these relevant attributes. The pipeline of the designed hierarchical strategy is depicted in Fig.\ref{fig:overview}(b), with detail provided in the Fig.11 of Appendix .

To effectively utilize reflective-metallic material information in the PBR diffusion model process, we inject material caption features into the U-Net. Specifically, we first use the CLIP text encoder~\cite{radford2021learning} to extract language features from the material captions. Then, to inject the conditioning signal into the Albedo U-Net network $\epsilon_{\theta}$ and the Metallic-Roughness U-Net network $\epsilon_{\sigma}$, we capture the embedding relationship using cross-attention mechanism between the material caption language features and the noised latent of the U-Net. We reparameterize the loss functions of $\epsilon_{\theta}^{a}$ and $\epsilon_{\theta}^{mr}$ following the corresponding conditioning signals in Eq.\ref{eq:3} and Eq.\ref{eq:4}:
\begin{equation}
\vspace{0.1cm}
\small
    L=\mathbb{E}_{\mathcal{E}(x),\mathcal{E}(I),\epsilon, t}
    [|\epsilon-\epsilon_{\theta}^{mr}(z_{t}^{mr}, t,f_{s}(\mathcal{E}(\bm{I})), C_T(\bm{I}))|_{2}^{2}]\label{eq:3}
\end{equation}
\begin{equation}
\small
    L=\mathbb{E}_{\mathcal{E}(x),\mathcal{E}(I),\epsilon, t}[|\epsilon-\epsilon_{\theta}^{a}(z_{t}^{a}, t, C_T(\bm{I})))|_{2}^{2}]\label{eq:4}
\end{equation}
\vspace{0.1cm}
\noindent where $C_T$ denotes CLIP text feature encoder, $z_{t}^{a}$ is the latent feature encoded from the ground truth Albedo at timestep $t$.

\subsubsection{View-dependent illumination-aware condition}\label{sec:SV-Illum}
Although the designed VLM-guided material diffusion model can effectively handle part-wise properties by injecting material description information, Metallic-Roughness estimation still struggles to capture spatially varying properties, due to the challenges posed by the severe long-tail distribution effect in the training data. Therefore, it is essential to introduce additional physics-based prior information.

Physically Based Rendering (PBR) materials often utilize the spatially-varying bi-directional reflectance distribution functions (svBRDFs) to approximate the surface reflectance property with a set of decomposed intrinsic terms, which can encapsulate the interaction of light with the object surface. Following the popular specular svBRDFs model, like the Cook-Torrance model~\cite{cook1982reflectance}, we can observe that the Roughness and Metallic properties are closely integrated with the specular shading term. The detailed rendering equation is provided in the Appendix (Sec.A.1). Specifically, the rendering equation is rewritten as Eq.~\ref{eq:render_equation_2}:
\begin{equation}
\vspace{-0.5cm}
\small
\begin{split}
    L_o(\hat{\textbf{x}},\bm{\omega}_o) &= L_{diff}(\hat{\textbf{x}}, k_{d}, L_i) \\
    & + L_{spec}(\hat{\textbf{x}}, \bm{\omega}_o, F_{0}(k_{m}), k_{r}, L_i) \\
    \label{eq:render_equation_2}
\end{split}
\vspace{-0.3cm}
\end{equation}
\noindent where the rendering results comprise diffuse shading $L_{diff}$ and specular shading $L_{spec}$, $k_{d}$ and $k_{r}$ represent Albedo and Roughness, $k_{m}$ is metallic term related to the Fresnel term. 

Previous methods~\cite{boss2024sf3d, zhang2024clay, siddiqui2024meta} for generating Roughness and Metallic maps tend to homogenize the distribution of BRDF in synthetic 3D data. For example, in the widely used Objaverse dataset~\cite{deitke2023objaverse}, we observe that the Metalness and Roughness of objects often have fixed values. This results in a lack of spatial variability in the BRDF distribution obtained through diffusion model training, which negatively impacts the accuracy of the Metalness and Roughness attributes. To overcome the severe impact of the long-tail distribution, and motivated by Eq.\ref{eq:render_equation_2}, we propose introducing view-dependent illumination as a condition to guide the sampling process of Roughness and Metallic. We further reformulate the loss function shown in Eq.\ref{eq:1} into the following form in Eq.~\ref{eq:2}:
\begin{equation}
\small
    L = \mathbb{E}_{\mathcal{E}(x),\mathcal{E}(I),\epsilon, t}[|\epsilon-\epsilon_{\theta}(z_{t}^{mr}, t,f_{s}(\mathcal{E}(\bm{I})))|_{2}^{2}]\label{eq:2}
\end{equation}
\noindent where $z_{t}^{mr}$ is the latent feature encoded from the ground truth Metallic-Roughness gt at time step $t$, $f_{s}$ is a specular illumination diffusion model to generate spatially-varying illumination based on single RGB grid image $\bm{I}$. This specular illumination map acts as a local cue, enhancing the Metallic and Roughness spatial distribution closely linked to the actual distribution of the object.

\subsection{Dual-head PBR Reconstruction Model}
Our PBR reconstruction model takes multi-view Albedo multi-view images and MRO images as input, and it outputs 3D assets with PBR material. Since our reconstruction model has two types of inputs, using a single image encoder to extract features from both types would cause the model to be unable to distinguish between them. Therefore, we use dual-head PBR encoders: one encoder is for encoding the Albedo images, and the other encoder is for encoding the MRO images. Then two kinds of image tokens and learnable triplane tokens are passed through a transformer backbone, outputting a low-resolution triplane. The low-resolution triplane has a resolution of 64 with 1024 channels. We use an upsampler similar to \cite{yang2024hunyuan3D} to upsample the low-resolution triplane, ultimately obtaining a high-resolution triplane with a resolution of 256 and 120 channels. For each query point, we project it onto a triplane to get its triplane feature, and we use an MLP to predict its signed distance, Albedo and MRO. 

Since NeRF\cite{mildenhall2021nerf} and 3D Gaussian\cite{kerbl20233D} have difficulty generating high-quality meshes, and the training of DMTet\cite{shen2021deep} and Flexicubes\cite{shen2023flexible} is unstable, we adopt NeuS\cite{wang2021neus} as our 3D representation and obtain the geometry through marching cubes. With geometry, Albedo, and MRO, we obtain 3D asset with PBR as our final output. During training, we apply image loss for both Albedo images and MRO images:
\begin{equation}
\begin{split}
L =& \sum_{i}{||{I_{i, A}}-I_{i, A}^{gt} ||}_{2}^{2} \\
      +&\sum_{i}{||{I_{i, MRO}}-I_{i, MRO}^{gt} ||}_{2}^{2} \\
      +& \lambda_{lpips}\sum_{i}{L_{lpips}({I_{i, A}}, I_{i, A}^{gt})} \\
      +& \lambda_{lpips}\sum_{i}{L_{lpips}({I_{i, MRO}}, I_{i, MRO}^{gt})} \\
      +& \lambda_{mask}\sum_{i}{||{M_{i}}-M_{i}^{gt} ||}_{2}^{2}
\end{split}
\end{equation}
\vspace{-0.0cm}
\noindent where $I_{i, A}$, ${I_{i, MRO}}$, ${M_{i}}$, $I_{i, A}^{gt}$, $I_{i, MRO}^{gt}$ and $M_{i}^{gt}$ denote rendered Albedo images, rendered MRO images, rendered mask images, ground truth Albedo images, ground truth MRO images, ground truth mask images of the $i$-th view. We randomly select 2 target views and set $\lambda_{lpips}=2$, $\lambda_{mask}=0.2$ during the training stage.

\begin{figure*}[thbp]
\vspace{-0.2cm}
\centering
  \includegraphics[width=0.88\linewidth]{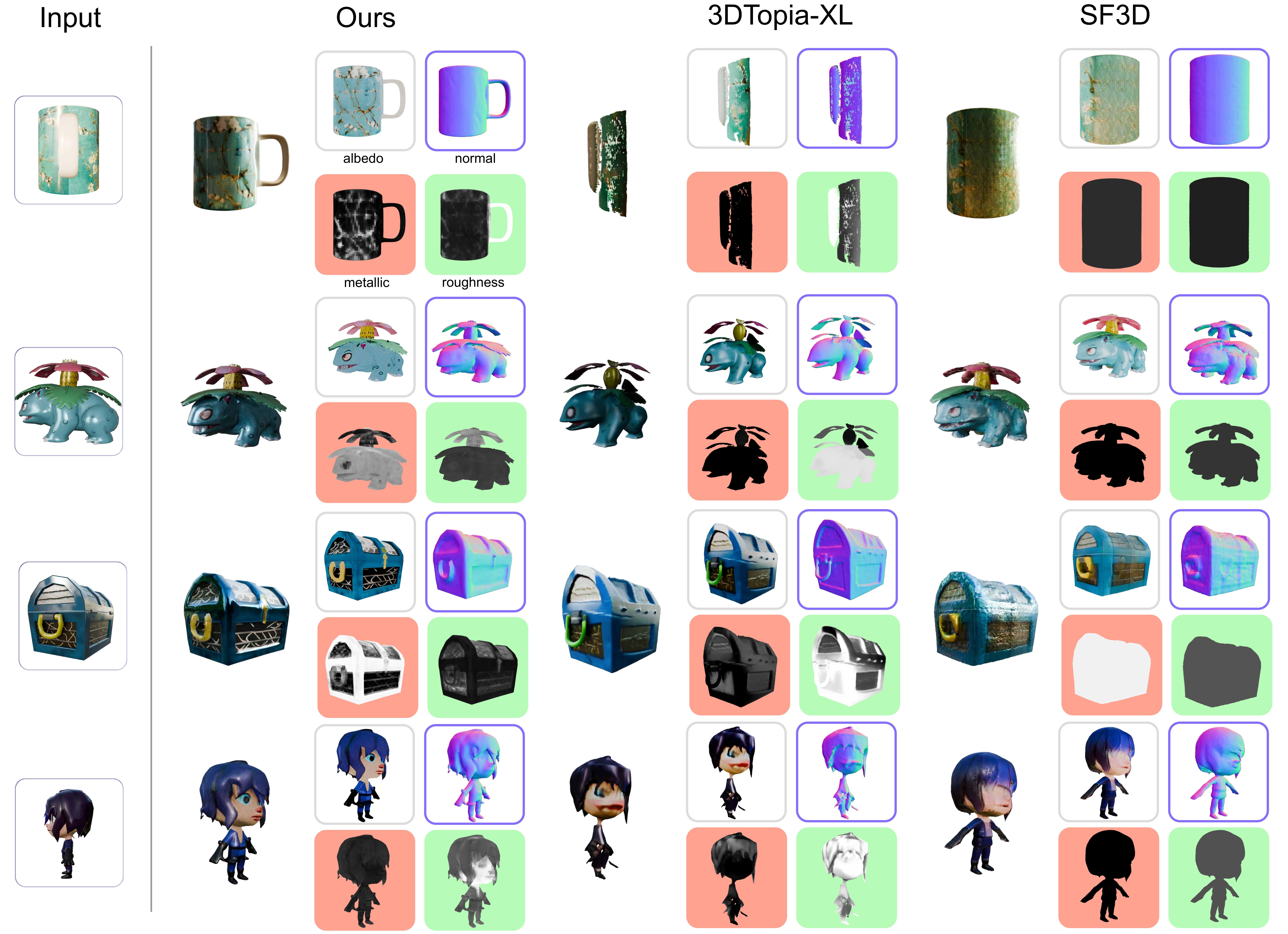}
  \vspace{-0.4cm}
\caption{ Qualitative comparison of the generated 3D assets with other methods. }
\label{fig:LRM_Comparison_image}
\vspace{-0.2cm}
\end{figure*}

\section{Experiments}

In this section, we evaluate PBR3DGen, detailing the experiment settings and datasets in Sec.\ref{sec:exp_set}, and compare it with SOTAs in 3D generation and albedo estimation based on diffusion models across various datasets in Sec.\ref{sec:compare}. Finally, we demonstrate the effectiveness of different components in improving PBR estimation and mesh generation in Sec.~\ref{sec:ablation}.
More results are shown in the Appendix.

\begin{table*}[htbp]  
\centering
\resizebox{1.\linewidth}{!}{
  \begin{tabular}{l|cc|cc|cc|cc|cccc}
    \toprule Methods
     &  \multicolumn{2}{c}{Albedo}  & \multicolumn{2}{c}{Roughness}  & \multicolumn{2}{c}{Metallic}  & \multicolumn{2}{c} {RGB} & \multicolumn{4}{c} {Geometry}\\
    & PSNR$\uparrow$ & MSE $\downarrow$ & PSNR $\uparrow$ & MSE $\downarrow$ & PSNR $\uparrow$ & MSE $\downarrow$ & PSNR $\uparrow$ & MSE $\downarrow$ & CD$\downarrow$ & FS@0.1$\uparrow$ & FS@0.2$\uparrow$ & FS@0.5$\uparrow$\\
    \midrule
    SF3D~\cite{boss2024sf3d} & 15.98 &0.030 & 15.00 & 0.040 &  16.04 & 0.035 & 16.37 & 0.028 & 0.332 & 0.607 & 0.787 & 0.919\\
    3DTopia-XL~\cite{chen20243dtopia} & 13.69 & 0.047 & 11.84 & 0.075 &  13.14 & 0.058 & 13.95 & 0.046 & 0.552 & 0.328 & 0.549 & 0.826\\
    \textbf{ Ours} & \textbf{17.77} & \textbf{0.022} & \textbf{16.52} & \textbf{0.030} &  \textbf{16.16} & \textbf{0.035} & \textbf{17.99} & \textbf{0.021} &\textbf{0.214} &\textbf{0.720} &\textbf{0.875} &\textbf{0.967}\\
    \bottomrule
  \end{tabular}
}
\vspace{-0.3cm}
\caption{Quantitative comparison on Objaverse~\cite{deitke2023objaverse} dataset.}
\label{tab:comparison_on_objaverse}
\vspace{-0.4cm}
\end{table*}

\begin{table}[htbp]  
\centering
\resizebox{1.\linewidth}{!}{
  \begin{tabular}{l|cccc}
    \toprule Methods
    & CD$\downarrow$ & FS@0.1$\uparrow$ & FS@0.2$\uparrow$ & FS@0.5$\uparrow$\\
    \midrule
    LGM \cite{tang2025lgm} & 0.409 & 0.442 & 0.658 & 0.881\\
    OpenLRM \cite{he2023openlrm} & 0.214 & 0.605 & 0.840 & \textbf{0.997}\\
    TripoSR~\cite{tochilkin2024triposr} & 0.356 & 0.511 & 0.727 & 0.920\\
    InstantMesh~\cite{xu2024instantmesh} & 0.216 & 0.670 & 0.862 & 0.977\\
    SF3D~\cite{boss2024sf3d} & 0.274 & 0.554 & 0.786 & 0.956\\
    3DTopia-XL~\cite{chen20243dtopia} & 0.239 & 0.635 & 0.832 & 0.962\\
    \textbf{Ours} & \textbf{0.175} & \textbf{0.739} & \textbf{0.903} & 0.984 \\
    \bottomrule
  \end{tabular}
}
\vspace{-0.3cm}
\caption{Quantitative comparison on GSO~\cite{downs2022google} dataset.}
\label{tab:exp_syn}
\vspace{-0.4cm}
\end{table}
\begin{figure}[htbp]
\centering
  \includegraphics[width=0.95\linewidth]{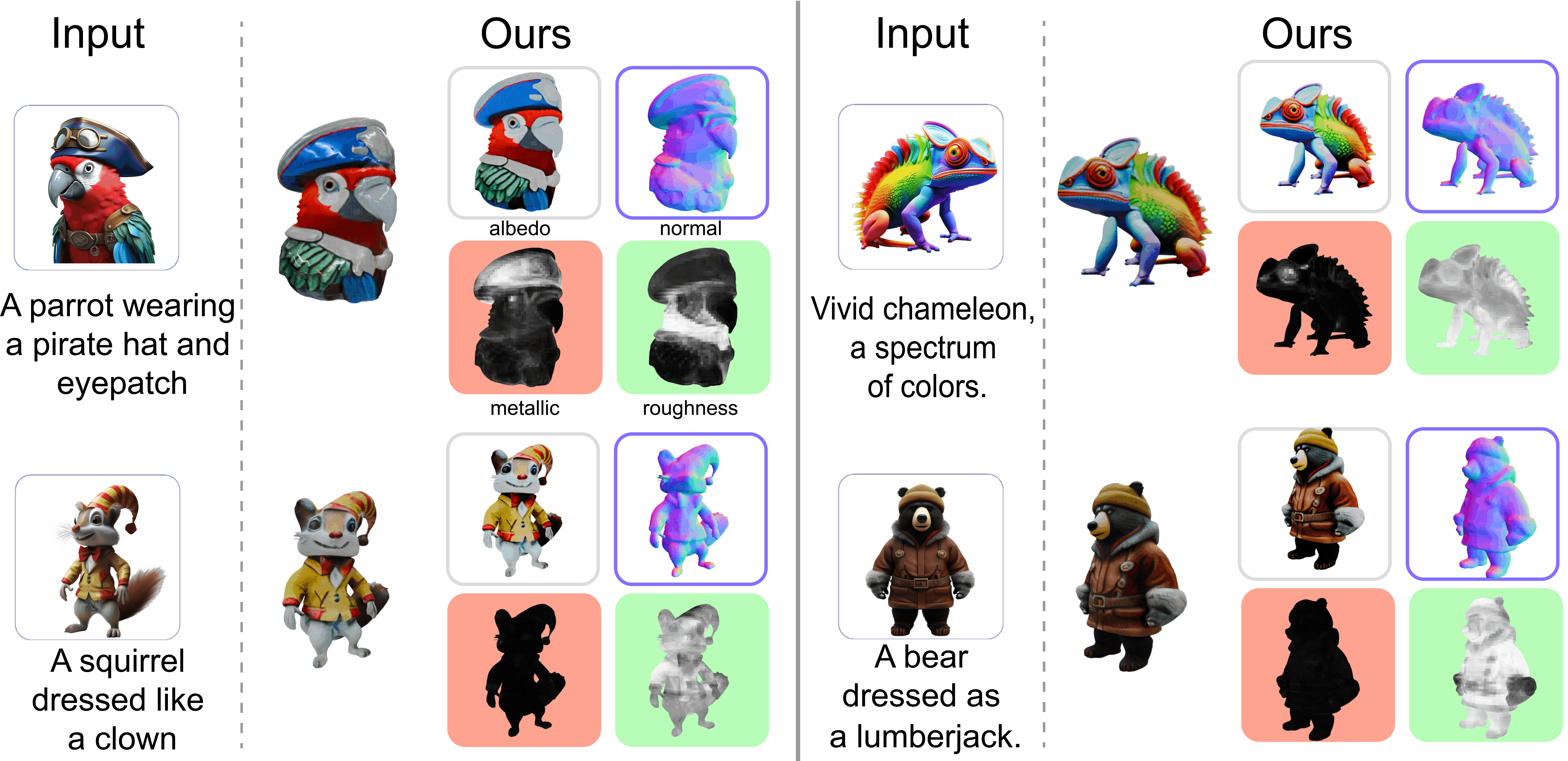}
  \vspace{-0.3cm}
\caption{ Text/Image to 3D results of our method. }
\label{fig:LRM_Comparison_text2mesh}
\vspace{-0.2cm}
\end{figure}

\subsection{Experimental Setup}\label{sec:exp_set}
\noindent \textbf{Dataset}. We first generate multi-view PBR material training data from the Objaverse dataset~\cite{deitke2023objaverse}. Specifically, we select 48k objects by filtering out those with low-quality PBR textures, including items with black or white albedo, as well as objects missing any essential PBR attributes. For each object, we render 21-view multi-domain images (RGB, Albedo, Metallic, Roughness, Specular light) in an orbiting trajectory with uniform azimuths and varying elevations in ${30^{\circ}, 0^{\circ}, -30^{\circ}}$, under one randomly selected HDR environment map from 120 HDRI maps. The entire dataset is rendered using Blender. To evaluate the reconstruction quality of PBR materials, we randomly select 300 objects with PBR from Objaverse~\cite{deitke2023objaverse}, due to the lack of publicly available PBR datasets. Additionally, we assess the quantitative performance of geometry and novel view synthesis using other public datasets, such as Google Scanned Objects (GSO)~\cite{downs2022google}.

\noindent \textbf{Implementation}. For PBR estimation model, we use a diffusion model fine-tuned from the Stable Diffusion V2.1~\cite{rombach2022high} based on the framework of Intructpix2pix~\cite{brooks2023instructpix2pix} as a starting point and continue to train for 50k iterations using an Adam~\cite{kingma2014adam} optimizer at a learning rate of \(1 \times 10^{-4}\). For the PBR reconstruction model, we pretrain the model with only the Albedo input for 200k iterations, and then we finetune the model for 150k iterations with both Albedo and MRO image input, the MRO image encoder is initialized with the weights of the Albedo image encoder. The learning rate is \(3 \times 10^{-5}\) in both stages.

\noindent \textbf{Metrics}. We evaluate both the 2D visual quality and 3D geometric quality of the generated assets. Specifically, we asses the image quality using PSNR and MSE between the rendered and ground-truth images in both of two phases, including multi-view PBR generation and mesh generation, and we compute in the foreground region to avoid metric inflation due to the empty background. 
For 3D geometric evaluation, we first align the coordinate system of the generated meshes with the ground-truth meshes and report Chamfer Distance (CD) and F-Score (FS), which are computed by sampling 10K points from the surface uniformly.

\subsection{Comparison with Other Methods}\label{sec:compare}
\noindent \textbf{Baselines}. We compare our results with state-of-the-art methods for both stages of our pipeline. For the multi-view PBR image generation stage, we additionally compare with IntrinsicAnything~\cite{chen2024intrinsicanything}, a diffusion-based intrinsic image generation method. For 3D asset generation methods, we mainly focus on 3D with PBR generation models to maintain a consistent evaluation protocol across different techniques. Specifically, we compare against SF3D~\cite{boss2024sf3d}, 3DTopia-XL~\cite{chen20243dtopia}, Instantmesh~\cite{xu2024instantmesh}, OpenLRM~\cite{he2023openlrm}, LGM~\cite{tang2025lgm}, TripoSR~\cite{tochilkin2024triposr}. 

\begin{figure}[htbp]
\centering
  \includegraphics[width=0.95\linewidth]{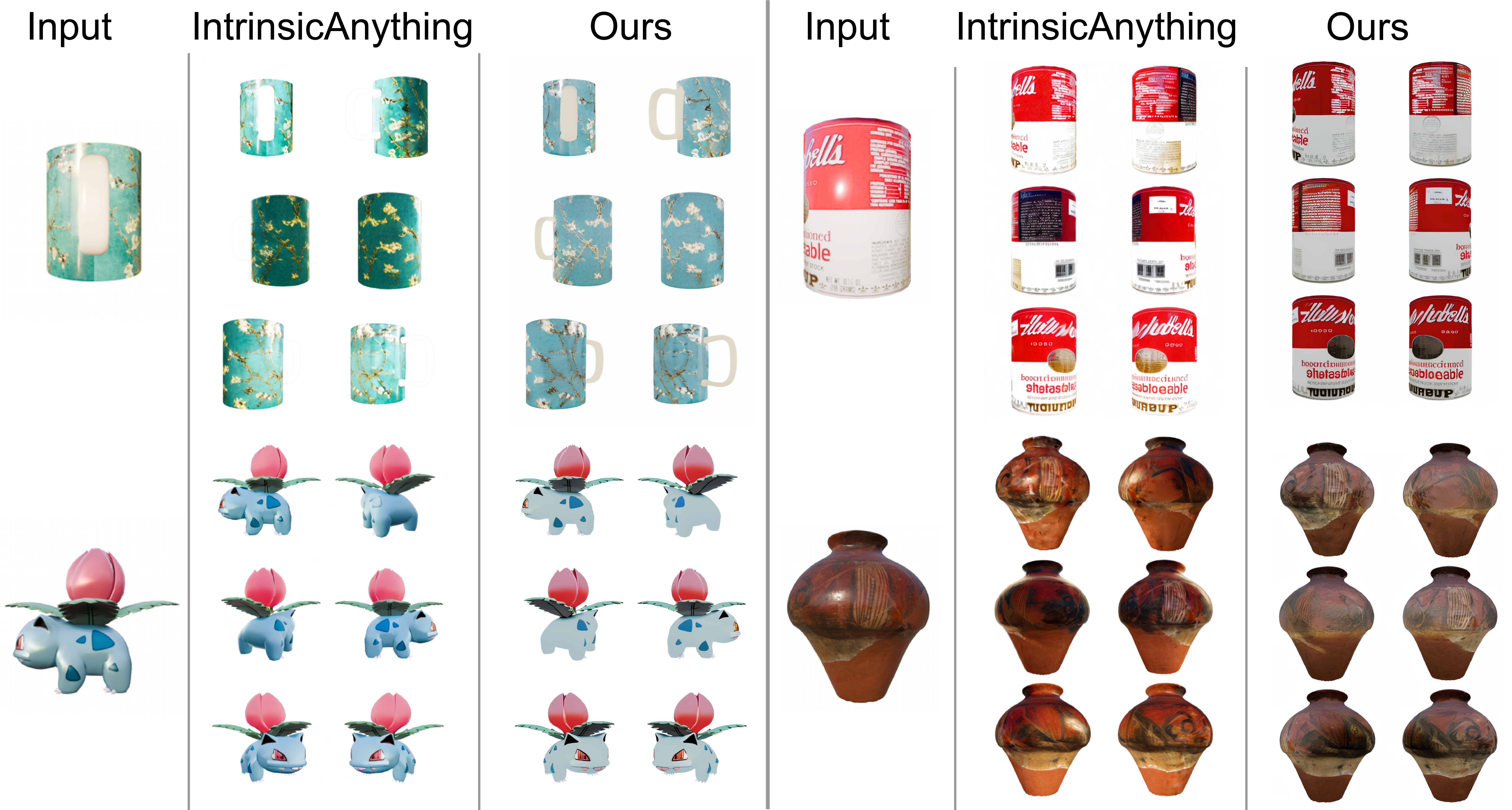}
  \vspace{-0.1cm}
\caption{ Qualitative comparison with IntrinsicAnything~\cite{chen2024intrinsicanything}. }
\label{fig:albedo_Comparison}
\vspace{-0.2cm}
\end{figure}

\noindent \textbf{Results}. From the 2D PBR image estimation metrics in Table~\ref{tab:ablation_albedo} and Fig~\ref{fig:albedo_Comparison}, we report our the quantitative and qualitative results outperforms IntrinsicAnything~\cite{chen2024intrinsicanything} on PSNR
and MSE significantly. 

Notably, experiments with IntrinsicAnything\cite{chen2024intrinsicanything} show an abundance of black albedos, especially for metallic objects, indicating its method's limited generalization to these materials. Nevertheless, we are able to significantly reduce the ambiguities related to these objects. Our method excels in handling highlights and shadows, particularly for metallic objects, which is essential for generating precise and reliable meshes with PBR materials in the large reconstruction module.

As shown in Table~\ref{tab:comparison_on_objaverse}, Table~\ref{tab:exp_syn}, our quantitative results outperform other baselines on novel view PBR assets and geometric metrics quality across all metrics. Given that SF3D~\cite{boss2024sf3d} and 3DTopia-XL~\cite{chen20243dtopia} are the only recently open-sourced models capable of generating PBR assets, we compare our 2D novel view PBR materials and RGB generation results with these methods. The qualitative results for image-to-mesh and text-to-mesh generation are shown in Figs~\ref{fig:LRM_Comparison_image} and Figs~\ref{fig:LRM_Comparison_text2mesh}. Our approach demonstrates superior performance in reconstructing shapes with detailed geometry and high-fidelity PBR textures. Thanks to the high-quality spatially-varying albedo and accurate metallic and roughness, free from any baked lighting and shadows, our relighting experiments exhibit promising results compared with other baselines. Notably, although our dataset contains fewer samples, the fine-tuning of the multi-view diffusion model with this data enables us to generate high-quality multi-view images and 3D assets with PBR.


\vspace{-0.15cm}
\begin{table}[htbp]  
\centering
\tiny
\resizebox{0.8\linewidth}{!}{
  \begin{tabular}{l|cc}
    \toprule Methods         
     & \multicolumn{2}{c}{Albedo}\\
    & PSNR$\uparrow$  & MSE $\downarrow$ 
 \\
    \midrule
    IntrinsicAnything~\cite{chen2024intrinsicanything} & 16.775 & 0.031 \\
    \textbf{\textbf{Ours}} & \textbf{18.186} & \textbf{0.023} \\
    \midrule
    w/o VLM-guide &  17.155 & 0.030 \\
    \bottomrule
  \end{tabular}
}
\vspace{-0.1cm}
\caption{Quantitative results for IntrinsicAnything~\cite{chen2024intrinsicanything} and ours. All scores are calculated as an average across 300 objects from the Objaverse dataset.}
\label{tab:ablation_albedo}
\vspace{-0.2cm}
\end{table}

\begin{table}[htbp]  
\centering
\resizebox{0.9\linewidth}{!}{
  \begin{tabular}{l|cc|cc}
    \toprule Methods
     & \multicolumn{2}{c}{Roughness} & \multicolumn{2}{c}{Metallic}\\
    & PSNR$\uparrow$  & MSE $\downarrow$ & PSNR$\uparrow$ & MSE $\downarrow$\\
    \midrule
    \makecell{w/o VD-Illum \\  w/o VLM-guide} & 18.649 & 0.020 & 15.293 & 0.0456\\
    \midrule
    w/o VD-Illum & 19.803 & 0.019 & 15.871 & 0.038\\
    \midrule
    w/o VLM-guide & 18.023 & 0.023 & 15.890 & 0.038\\
    \midrule
    \textbf{Ours} & \textbf{21.095} & \textbf{0.013} & \textbf{17.718} & \textbf{0.028}\\
    \bottomrule
  \end{tabular}
}
\vspace{-0.3cm}
\caption{Ablation studies for Mulit-view PBR model.  All scores are calculated as an average across 300 objects from the Objaverse dataset.}
\label{tab:ablation_mro}
\vspace{-0.45cm}
\end{table}

\subsection{Ablation Study}\label{sec:ablation}
\begin{figure}[htbp]
\centering
  \includegraphics[width=1\linewidth]{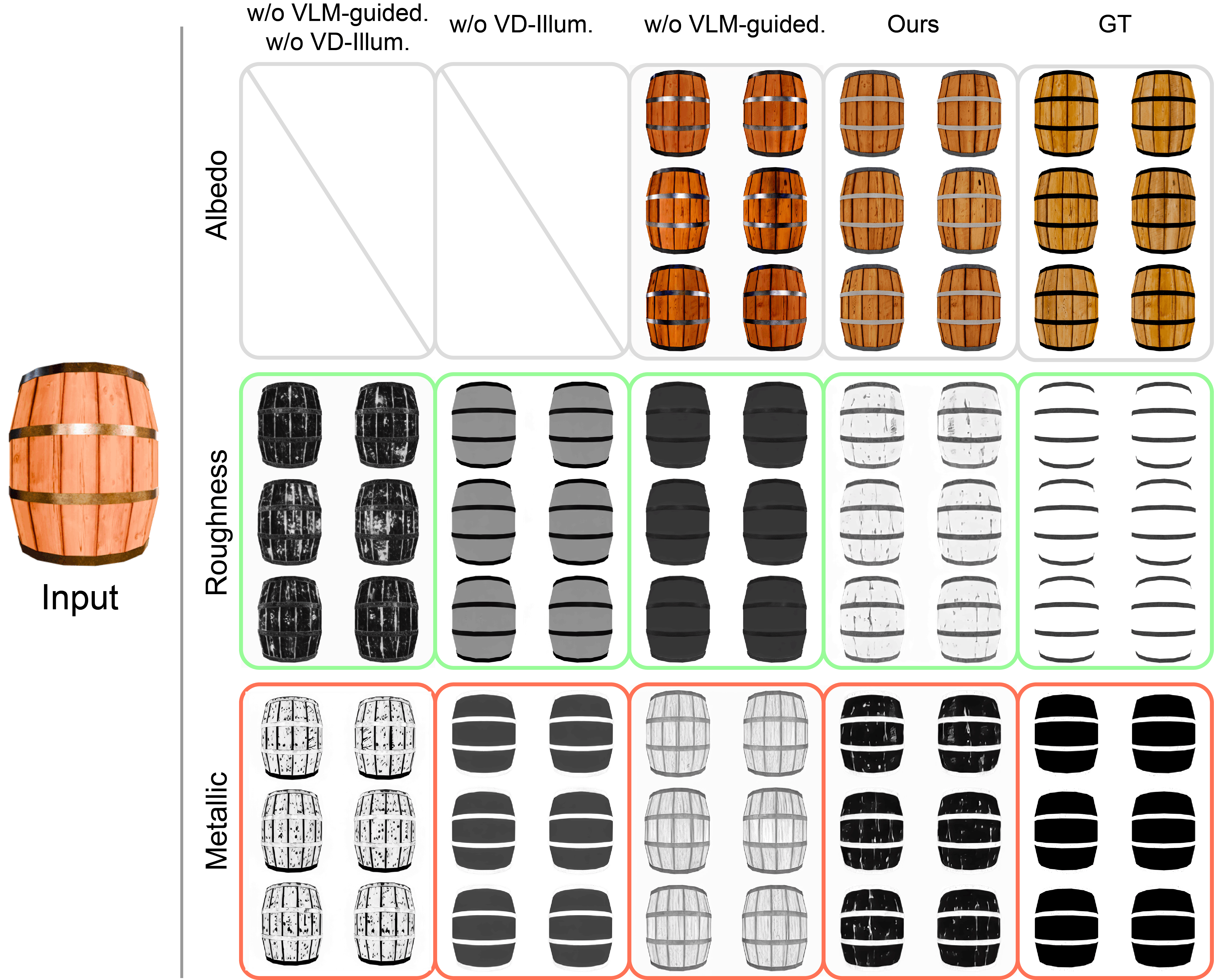}
  \vspace{-0.3cm}
\caption{ Qualitative ablation on PBR estimation. }
\label{fig:pbr_ablation}
\vspace{-0.2cm}
\end{figure}

We conduct ablation studies to analyze the contribution of each component in our framework on the Objaverse PBR dataset. The key innovation of our method is its ability to effectively estimate multi-view spatially varying PBR materials. Therefore, this section focuses on evaluating the impact of the VLM-guided PBR material generation module in Sec.~\ref{sec:VLM-guided} and the view-dependent illumination-aware condition module in Sec.~\ref{sec:SV-Illum}. Quantitative and qualitative results are presented in Table~\ref{tab:ablation_albedo}, Table~\ref{tab:ablation_mro}, Fig~\ref{fig:pbr_ablation} and  Fig~\ref{fig:albedo_Comparison}, respectively.

\noindent\textbf{(1) w/o VD-Illum.} This term indicates training without the view-dependent illumination condition for the multi-view metallic-roughness diffusion model.
As shown in Table~\ref{tab:ablation_mro} and Fig~\ref{fig:pbr_ablation}, VD-Illum. module as local pixel-aware priors prove to be effective for boosting the metallic-roughness distribution according to spatially varying attribution.

\noindent\textbf{(2) w/o VLM-guide.} This term indicates training without the VLM-guided condition signal for both multi-view albedo and metallic-roughness diffusion models. This is to demonstrate the necessity of introducing VLM-guided conditions for accurate PBR estimation. Table~\ref{tab:ablation_mro} clearly shows that removing the VLM-guided module leads to a remarkable decline in the performance of metallic-roughness accuracy, adversely affecting the albedo estimation as well. The qualitative results in Fig~\ref{fig:pbr_ablation} further indicate the absence of VLM-guided module results in incorrect albedo outputs with baked highlight. Simultaneously, this module plays a pivotal role in rectifying the distribution of predicted roughness and metallic values. By integrating VLM material priors for metalness and reflectivity into our metallic and roughness diffusion model, which considers materials from both global and local priors, we significantly reduce the discrepancy between predicted and actual distributions, a problem often caused by homogeneous training data.

\noindent\textbf{(3) w/o VD-Illum.\& VLM-guide.} This ablation is to evaluate the multi-view PBR diffusion model without both of the above conditions. From the first column of Fig~\ref{fig:pbr_ablation}, it is evident that without the VLM-guided condition and the VD-illum condition, the predicted distributions for roughness and metallic are more disorganized. Additionally, the predicted values for roughness deviate significantly from the ground truth roughness image.

\subsection{Limitations and Future Work}
Despite the effectiveness of our PBR3DGen, it achieves promising 3D mesh with PBR materials quality, yet it has limitations. Specifically, our framework relies on the RGB Multi-view diffusion model, which is limited by the capabilities of the Multi-View diffusion model. Poor geometric consistency can degrade the quality of our subsequent reconstruction models. In future work, we aim to improve the reconstruction of specular and transparent objects. Current diffusion models struggle to handle the geometric consistency of such objects. To address this, we plan to develop a physics-based prior that uses principles of specular reflections, refractions, and surfaces. This will enhance both the accuracy and quality of reconstructed geometry and appearance for these challenging materials.

\section{Conclusion}
We present PBR3DGen, a two-stage 3D mesh with high-fidelity PBR materials generation framework by introducing a novel PBR multi-view diffusion model. Specifically, we explore unleashing VLM strong material prior and view-dependent illumination-aware conditions to guide PBR multi-view generation, which significantly alleviates the ambiguity often encountered between highlights and albedo within the rendered imagery and handles spatially-varying properties. Thanks to accurate PBR multi-view images, we significantly improved 3D mesh with PBR materials reconstruction quality based on our dual-head PBR reconstruction model.

{
    \small
    \bibliographystyle{ieeenat_fullname}
    \bibliography{main}

\begin{thebibliography}{49}
\providecommand{\natexlab}[1]{#1}
\providecommand{\url}[1]{\texttt{#1}}
\expandafter\ifx\csname urlstyle\endcsname\relax
  \providecommand{\doi}[1]{doi: #1}\else
  \providecommand{\doi}{doi: \begingroup \urlstyle{rm}\Url}\fi

\bibitem[Agarwal et~al.(2011)Agarwal, Furukawa, Snavely, Simon, Curless, Seitz, and Szeliski]{agarwal2011building}
Sameer Agarwal, Yasutaka Furukawa, Noah Snavely, Ian Simon, Brian Curless, Steven~M Seitz, and Richard Szeliski.
\newblock Building rome in a day.
\newblock \emph{Communications of the ACM}, 54\penalty0 (10):\penalty0 105--112, 2011.

\bibitem[Boss et~al.(2024)Boss, Huang, Vasishta, and Jampani]{boss2024sf3d}
Mark Boss, Zixuan Huang, Aaryaman Vasishta, and Varun Jampani.
\newblock Sf3d: Stable fast 3d mesh reconstruction with uv-unwrapping and illumination disentanglement.
\newblock \emph{arXiv preprint arXiv:2408.00653}, 2024.

\bibitem[Brooks et~al.(2023)Brooks, Holynski, and Efros]{brooks2023instructpix2pix}
Tim Brooks, Aleksander Holynski, and Alexei~A Efros.
\newblock Instructpix2pix: Learning to follow image editing instructions.
\newblock In \emph{Proceedings of the IEEE/CVF Conference on Computer Vision and Pattern Recognition}, 2023.

\bibitem[Cao et~al.(2023)Cao, Kreis, Fidler, Sharp, and Yin]{cao2023texfusion}
Tianshi Cao, Karsten Kreis, Sanja Fidler, Nicholas Sharp, and Kangxue Yin.
\newblock Texfusion: Synthesizing 3d textures with text-guided image diffusion models.
\newblock In \emph{Proceedings of the IEEE/CVF International Conference on Computer Vision}, 2023.

\bibitem[Chen et~al.(2023{\natexlab{a}})Chen, Siddiqui, Lee, Tulyakov, and Nie{\ss}ner]{chen2023text2tex}
Dave~Zhenyu Chen, Yawar Siddiqui, Hsin-Ying Lee, Sergey Tulyakov, and Matthias Nie{\ss}ner.
\newblock Text2tex: Text-driven texture synthesis via diffusion models.
\newblock In \emph{Proceedings of the IEEE/CVF International Conference on Computer Vision}, 2023{\natexlab{a}}.

\bibitem[Chen et~al.(2023{\natexlab{b}})Chen, Chen, Jiao, and Jia]{chen2023fantasia3d}
Rui Chen, Yongwei Chen, Ningxin Jiao, and Kui Jia.
\newblock Fantasia3d: Disentangling geometry and appearance for high-quality text-to-3d content creation.
\newblock In \emph{Proceedings of the IEEE/CVF international conference on computer vision}, 2023{\natexlab{b}}.

\bibitem[Chen et~al.(2024{\natexlab{a}})Chen, Peng, Yang, Liu, Pan, Lv, and Zhou]{chen2024intrinsicanything}
Xi Chen, Sida Peng, Dongchen Yang, Yuan Liu, Bowen Pan, Chengfei Lv, and Xiaowei Zhou.
\newblock Intrinsicanything: Learning diffusion priors for inverse rendering under unknown illumination.
\newblock \emph{arXiv preprint arXiv:2404.11593}, 2024{\natexlab{a}}.

\bibitem[Chen et~al.(2024{\natexlab{b}})Chen, Tang, Dong, Cao, Hong, Lan, Wang, Xie, Wu, Saito, et~al.]{chen20243dtopia}
Zhaoxi Chen, Jiaxiang Tang, Yuhao Dong, Ziang Cao, Fangzhou Hong, Yushi Lan, Tengfei Wang, Haozhe Xie, Tong Wu, Shunsuke Saito, et~al.
\newblock 3dtopia-xl: Scaling high-quality 3d asset generation via primitive diffusion.
\newblock \emph{arXiv preprint arXiv:2409.12957}, 2024{\natexlab{b}}.

\bibitem[Cook and Torrance(1982)]{cook1982reflectance}
Robert~L Cook and Kenneth~E. Torrance.
\newblock A reflectance model for computer graphics.
\newblock \emph{ACM Transactions on Graphics (ToG)}, 1\penalty0 (1), 1982.

\bibitem[Deitke et~al.(2023)Deitke, Schwenk, Salvador, Weihs, Michel, VanderBilt, Schmidt, Ehsani, Kembhavi, and Farhadi]{deitke2023objaverse}
Matt Deitke, Dustin Schwenk, Jordi Salvador, Luca Weihs, Oscar Michel, Eli VanderBilt, Ludwig Schmidt, Kiana Ehsani, Aniruddha Kembhavi, and Ali Farhadi.
\newblock Objaverse: A universe of annotated 3d objects.
\newblock In \emph{Proceedings of the IEEE/CVF Conference on Computer Vision and Pattern Recognition}, 2023.

\bibitem[Deng et~al.(2024)Deng, Omernick, Weiss, Ramanan, Zhu, Zhou, and Agrawala]{deng2024flashtex}
Kangle Deng, Timothy Omernick, Alexander Weiss, Deva Ramanan, Jun-Yan Zhu, Tinghui Zhou, and Maneesh Agrawala.
\newblock Flashtex: Fast relightable mesh texturing with lightcontrolnet.
\newblock \emph{arXiv preprint arXiv:2402.13251}, 2024.

\bibitem[Downs et~al.(2022)Downs, Francis, Koenig, Kinman, Hickman, Reymann, McHugh, and Vanhoucke]{downs2022google}
Laura Downs, Anthony Francis, Nate Koenig, Brandon Kinman, Ryan Hickman, Krista Reymann, Thomas~B McHugh, and Vincent Vanhoucke.
\newblock Google scanned objects: A high-quality dataset of 3d scanned household items.
\newblock In \emph{2022 International Conference on Robotics and Automation (ICRA)}. IEEE, 2022.

\bibitem[Fang et~al.(2024)Fang, Sun, Wu, Wang, Liu, Wetzstein, and Lin]{fang2024make}
Ye Fang, Zeyi Sun, Tong Wu, Jiaqi Wang, Ziwei Liu, Gordon Wetzstein, and Dahua Lin.
\newblock Make-it-real: Unleashing large multimodal model for painting 3d objects with realistic materials.
\newblock \emph{arXiv preprint arXiv:2404.16829}, 3, 2024.

\bibitem[Furukawa and Ponce(2009)]{furukawa2009accurate}
Yasutaka Furukawa and Jean Ponce.
\newblock Accurate, dense, and robust multiview stereopsis.
\newblock \emph{IEEE transactions on pattern analysis and machine intelligence}, 32\penalty0 (8):\penalty0 1362--1376, 2009.

\bibitem[He and Wang(2023)]{he2023openlrm}
Zexin He and Tengfei Wang.
\newblock Openlrm: Open-source large reconstruction models, 2023.

\bibitem[Hong et~al.(2023)Hong, Zhang, Gu, Bi, Zhou, Liu, Liu, Sunkavalli, Bui, and Tan]{hong2023lrm}
Yicong Hong, Kai Zhang, Jiuxiang Gu, Sai Bi, Yang Zhou, Difan Liu, Feng Liu, Kalyan Sunkavalli, Trung Bui, and Hao Tan.
\newblock Lrm: Large reconstruction model for single image to 3d.
\newblock \emph{arXiv preprint arXiv:2311.04400}, 2023.

\bibitem[Kerbl et~al.(2023)Kerbl, Kopanas, Leimk{\"u}hler, and Drettakis]{kerbl20233D}
Bernhard Kerbl, Georgios Kopanas, Thomas Leimk{\"u}hler, and George Drettakis.
\newblock 3d gaussian splatting for real-time radiance field rendering.
\newblock \emph{ACM Trans. Graph.}, 42\penalty0 (4):\penalty0 139--1, 2023.

\bibitem[Kingma(2014)]{kingma2014adam}
Diederik~P Kingma.
\newblock Adam: A method for stochastic optimization.
\newblock \emph{arXiv preprint arXiv:1412.6980}, 2014.

\bibitem[Li et~al.(2024)Li, Liu, Long, Zhang, Lin, Li, Qi, Zhang, Luo, Tan, et~al.]{li2024era3d}
Peng Li, Yuan Liu, Xiaoxiao Long, Feihu Zhang, Cheng Lin, Mengfei Li, Xingqun Qi, Shanghang Zhang, Wenhan Luo, Ping Tan, et~al.
\newblock Era3d: High-resolution multiview diffusion using efficient row-wise attention.
\newblock \emph{arXiv preprint arXiv:2405.11616}, 2024.

\bibitem[Li et~al.(2023)Li, Li, Zhu, Yu, Zhao, Wan, You, Shi, and Lin]{li2023instant}
Sixu Li, Chaojian Li, Wenbo Zhu, Boyang Yu, Yang Zhao, Cheng Wan, Haoran You, Huihong Shi, and Yingyan Lin.
\newblock Instant-3d: Instant neural radiance field training towards on-device ar/vr 3d reconstruction.
\newblock In \emph{Proceedings of the 50th Annual International Symposium on Computer Architecture}, pages 1--13, 2023.

\bibitem[Li et~al.(2022)Li, He, Jiang, Liu, Tao, and Hai]{li2022meshformer}
Yuan Li, Xiangyang He, Yankai Jiang, Huan Liu, Yubo Tao, and Lin Hai.
\newblock Meshformer: High-resolution mesh segmentation with graph transformer.
\newblock In \emph{Computer Graphics Forum}, pages 37--49. Wiley Online Library, 2022.

\bibitem[Liu et~al.(2023{\natexlab{a}})Liu, Lin, Zeng, Long, Liu, Komura, and Wang]{liu2023syncdreamer}
Yuan Liu, Cheng Lin, Zijiao Zeng, Xiaoxiao Long, Lingjie Liu, Taku Komura, and Wenping Wang.
\newblock Syncdreamer: Generating multiview-consistent images from a single-view image.
\newblock \emph{arXiv preprint arXiv:2309.03453}, 2023{\natexlab{a}}.

\bibitem[Liu et~al.(2023{\natexlab{b}})Liu, Xie, Liu, and Wong]{liu2023text}
Yuxin Liu, Minshan Xie, Hanyuan Liu, and Tien-Tsin Wong.
\newblock Text-guided texturing by synchronized multi-view diffusion.
\newblock \emph{arXiv preprint arXiv:2311.12891}, 2023{\natexlab{b}}.

\bibitem[Long et~al.(2024)Long, Guo, Lin, Liu, Dou, Liu, Ma, Zhang, Habermann, Theobalt, et~al.]{long2024wonder3d}
Xiaoxiao Long, Yuan-Chen Guo, Cheng Lin, Yuan Liu, Zhiyang Dou, Lingjie Liu, Yuexin Ma, Song-Hai Zhang, Marc Habermann, Christian Theobalt, et~al.
\newblock Wonder3d: Single image to 3d using cross-domain diffusion.
\newblock In \emph{Proceedings of the IEEE/CVF Conference on Computer Vision and Pattern Recognition}, pages 9970--9980, 2024.

\bibitem[Mildenhall et~al.(2021)Mildenhall, Srinivasan, Tancik, Barron, Ramamoorthi, and Ng]{mildenhall2021nerf}
Ben Mildenhall, Pratul~P Srinivasan, Matthew Tancik, Jonathan~T Barron, Ravi Ramamoorthi, and Ren Ng.
\newblock Nerf: Representing scenes as neural radiance fields for view synthesis.
\newblock \emph{Communications of the ACM}, 65\penalty0 (1):\penalty0 99--106, 2021.

\bibitem[Pollefeys et~al.(2004)Pollefeys, Van~Gool, Vergauwen, Verbiest, Cornelis, Tops, and Koch]{pollefeys2004visual}
Marc Pollefeys, Luc Van~Gool, Maarten Vergauwen, Frank Verbiest, Kurt Cornelis, Jan Tops, and Reinhard Koch.
\newblock Visual modeling with a hand-held camera.
\newblock \emph{International Journal of Computer Vision}, 59:\penalty0 207--232, 2004.

\bibitem[Pollefeys et~al.(2008)Pollefeys, Nist{\'e}r, Frahm, Akbarzadeh, Mordohai, Clipp, Engels, Gallup, Kim, Merrell, et~al.]{pollefeys2008detailed}
Marc Pollefeys, David Nist{\'e}r, J-M Frahm, Amir Akbarzadeh, Philippos Mordohai, Brian Clipp, Chris Engels, David Gallup, S-J Kim, Paul Merrell, et~al.
\newblock Detailed real-time urban 3d reconstruction from video.
\newblock \emph{International Journal of Computer Vision}, 78:\penalty0 143--167, 2008.

\bibitem[Radford et~al.(2021)Radford, Kim, Hallacy, Ramesh, Goh, Agarwal, Sastry, Askell, Mishkin, Clark, et~al.]{radford2021learning}
Alec Radford, Jong~Wook Kim, Chris Hallacy, Aditya Ramesh, Gabriel Goh, Sandhini Agarwal, Girish Sastry, Amanda Askell, Pamela Mishkin, Jack Clark, et~al.
\newblock Learning transferable visual models from natural language supervision.
\newblock In \emph{International conference on machine learning}. PMLR, 2021.

\bibitem[Richardson et~al.(2023)Richardson, Metzer, Alaluf, Giryes, and Cohen-Or]{richardson2023texture}
Elad Richardson, Gal Metzer, Yuval Alaluf, Raja Giryes, and Daniel Cohen-Or.
\newblock Texture: Text-guided texturing of 3d shapes.
\newblock In \emph{ACM SIGGRAPH 2023 conference proceedings}, 2023.

\bibitem[Rombach et~al.(2022)Rombach, Blattmann, Lorenz, Esser, and Ommer]{rombach2022high}
Robin Rombach, Andreas Blattmann, Dominik Lorenz, Patrick Esser, and Bj{\"o}rn Ommer.
\newblock High-resolution image synthesis with latent diffusion models.
\newblock In \emph{Proceedings of the IEEE/CVF conference on computer vision and pattern recognition}, 2022.

\bibitem[Schonberger and Frahm(2016)]{schonberger2016structure}
Johannes~L Schonberger and Jan-Michael Frahm.
\newblock Structure-from-motion revisited.
\newblock In \emph{Proceedings of the IEEE conference on computer vision and pattern recognition}, pages 4104--4113, 2016.

\bibitem[Sch{\"o}nberger et~al.(2016)Sch{\"o}nberger, Zheng, Frahm, and Pollefeys]{schonberger2016pixelwise}
Johannes~L Sch{\"o}nberger, Enliang Zheng, Jan-Michael Frahm, and Marc Pollefeys.
\newblock Pixelwise view selection for unstructured multi-view stereo.
\newblock In \emph{Computer Vision--ECCV 2016: 14th European Conference, Amsterdam, The Netherlands, October 11-14, 2016, Proceedings, Part III 14}, pages 501--518. Springer, 2016.

\bibitem[Shen et~al.(2021)Shen, Gao, Yin, Liu, and Fidler]{shen2021deep}
Tianchang Shen, Jun Gao, Kangxue Yin, Ming-Yu Liu, and Sanja Fidler.
\newblock Deep marching tetrahedra: a hybrid representation for high-resolution 3d shape synthesis.
\newblock \emph{Advances in Neural Information Processing Systems}, 34:\penalty0 6087--6101, 2021.

\bibitem[Shen et~al.(2023)Shen, Munkberg, Hasselgren, Yin, Wang, Chen, Gojcic, Fidler, Sharp, and Gao]{shen2023flexible}
Tianchang Shen, Jacob Munkberg, Jon Hasselgren, Kangxue Yin, Zian Wang, Wenzheng Chen, Zan Gojcic, Sanja Fidler, Nicholas Sharp, and Jun Gao.
\newblock Flexible isosurface extraction for gradient-based mesh optimization.
\newblock \emph{ACM Trans. Graph.}, 42\penalty0 (4):\penalty0 37--1, 2023.

\bibitem[Shi et~al.(2023{\natexlab{a}})Shi, Chen, Zhang, Liu, Xu, Wei, Chen, Zeng, and Su]{shi2023zero123++}
Ruoxi Shi, Hansheng Chen, Zhuoyang Zhang, Minghua Liu, Chao Xu, Xinyue Wei, Linghao Chen, Chong Zeng, and Hao Su.
\newblock Zero123++: a single image to consistent multi-view diffusion base model.
\newblock \emph{arXiv preprint arXiv:2310.15110}, 2023{\natexlab{a}}.

\bibitem[Shi et~al.(2023{\natexlab{b}})Shi, Wang, Ye, Long, Li, and Yang]{shi2023mvdream}
Yichun Shi, Peng Wang, Jianglong Ye, Mai Long, Kejie Li, and Xiao Yang.
\newblock Mvdream: Multi-view diffusion for 3d generation.
\newblock \emph{arXiv preprint arXiv:2308.16512}, 2023{\natexlab{b}}.

\bibitem[Siddiqui et~al.(2024)Siddiqui, Monnier, Kokkinos, Kariya, Kleiman, Garreau, Gafni, Neverova, Vedaldi, Shapovalov, et~al.]{siddiqui2024meta}
Yawar Siddiqui, Tom Monnier, Filippos Kokkinos, Mahendra Kariya, Yanir Kleiman, Emilien Garreau, Oran Gafni, Natalia Neverova, Andrea Vedaldi, Roman Shapovalov, et~al.
\newblock Meta 3d assetgen: Text-to-mesh generation with high-quality geometry, texture, and pbr materials.
\newblock \emph{arXiv preprint arXiv:2407.02445}, 2024.

\bibitem[Tang et~al.(2025)Tang, Chen, Chen, Wang, Zeng, and Liu]{tang2025lgm}
Jiaxiang Tang, Zhaoxi Chen, Xiaokang Chen, Tengfei Wang, Gang Zeng, and Ziwei Liu.
\newblock Lgm: Large multi-view gaussian model for high-resolution 3d content creation.
\newblock In \emph{European Conference on Computer Vision}. Springer, 2025.

\bibitem[Tang et~al.(2023)Tang, Zhang, Chen, Wang, and Furukawa]{Tang2023mvdiffusion}
Shitao Tang, Fuyang Zhang, Jiacheng Chen, Peng Wang, and Yasutaka Furukawa.
\newblock Mvdiffusion: Enabling holistic multi-view image generation with correspondence-aware diffusion.
\newblock \emph{arXiv}, 2023.

\bibitem[Tochilkin et~al.(2024)Tochilkin, Pankratz, Liu, Huang, Letts, Li, Liang, Laforte, Jampani, and Cao]{tochilkin2024triposr}
Dmitry Tochilkin, David Pankratz, Zexiang Liu, Zixuan Huang, Adam Letts, Yangguang Li, Ding Liang, Christian Laforte, Varun Jampani, and Yan-Pei Cao.
\newblock Triposr: Fast 3d object reconstruction from a single image.
\newblock \emph{arXiv preprint arXiv:2403.02151}, 2024.

\bibitem[Wang et~al.(2021)Wang, Liu, Liu, Theobalt, Komura, and Wang]{wang2021neus}
Peng Wang, Lingjie Liu, Yuan Liu, Christian Theobalt, Taku Komura, and Wenping Wang.
\newblock Neus: Learning neural implicit surfaces by volume rendering for multi-view reconstruction.
\newblock \emph{arXiv preprint arXiv:2106.10689}, 2021.

\bibitem[Wei et~al.(2024)Wei, Zhang, Bi, Tan, Luan, Deschaintre, Sunkavalli, Su, and Xu]{wei2024meshlrm}
Xinyue Wei, Kai Zhang, Sai Bi, Hao Tan, Fujun Luan, Valentin Deschaintre, Kalyan Sunkavalli, Hao Su, and Zexiang Xu.
\newblock Meshlrm: Large reconstruction model for high-quality mesh.
\newblock \emph{arXiv preprint arXiv:2404.12385}, 2024.

\bibitem[Xu et~al.(2024)Xu, Cheng, Gao, Wang, Gao, and Shan]{xu2024instantmesh}
Jiale Xu, Weihao Cheng, Yiming Gao, Xintao Wang, Shenghua Gao, and Ying Shan.
\newblock Instantmesh: Efficient 3d mesh generation from a single image with sparse-view large reconstruction models.
\newblock \emph{arXiv preprint arXiv:2404.07191}, 2024.

\bibitem[Yang et~al.(2024)Yang, Shi, Zhang, Yang, Wang, Zhao, Liu, Wang, Lin, Yu, et~al.]{yang2024hunyuan3D}
Xianghui Yang, Huiwen Shi, Bowen Zhang, Fan Yang, Jiacheng Wang, Hongxu Zhao, Xinhai Liu, Xinzhou Wang, Qingxiang Lin, Jiaao Yu, et~al.
\newblock Hunyuan3d-1.0: A unified framework for text-to-3d and image-to-3d generation.
\newblock \emph{arXiv preprint arXiv:2411.02293}, 2024.

\bibitem[Zeng et~al.(2024{\natexlab{a}})Zeng, Chen, Qi, Liu, Zhao, Wang, Fu, Liu, and Yu]{zeng2024paint3d}
Xianfang Zeng, Xin Chen, Zhongqi Qi, Wen Liu, Zibo Zhao, Zhibin Wang, Bin Fu, Yong Liu, and Gang Yu.
\newblock Paint3d: Paint anything 3d with lighting-less texture diffusion models.
\newblock In \emph{Proceedings of the IEEE/CVF Conference on Computer Vision and Pattern Recognition}, pages 4252--4262, 2024{\natexlab{a}}.

\bibitem[Zeng et~al.(2024{\natexlab{b}})Zeng, Deschaintre, Georgiev, Hold-Geoffroy, Hu, Luan, Yan, and Ha{\v{s}}an]{zeng2024rgb}
Zheng Zeng, Valentin Deschaintre, Iliyan Georgiev, Yannick Hold-Geoffroy, Yiwei Hu, Fujun Luan, Ling-Qi Yan, and Milo{\v{s}} Ha{\v{s}}an.
\newblock Rgb\ensuremath{\leftrightarrow}x: Image decomposition and synthesis using material-and lighting-aware diffusion models.
\newblock In \emph{ACM SIGGRAPH 2024 Conference Papers}, 2024{\natexlab{b}}.

\bibitem[Zhang et~al.(2024{\natexlab{a}})Zhang, Song, Wei, Chen, Lu, and Tang]{zhang2024geolrm}
Chubin Zhang, Hongliang Song, Yi Wei, Yu Chen, Jiwen Lu, and Yansong Tang.
\newblock Geolrm: Geometry-aware large reconstruction model for high-quality 3d gaussian generation.
\newblock \emph{arXiv preprint arXiv:2406.15333}, 2024{\natexlab{a}}.

\bibitem[Zhang et~al.(2025)Zhang, Bi, Tan, Xiangli, Zhao, Sunkavalli, and Xu]{zhang2025gs}
Kai Zhang, Sai Bi, Hao Tan, Yuanbo Xiangli, Nanxuan Zhao, Kalyan Sunkavalli, and Zexiang Xu.
\newblock Gs-lrm: Large reconstruction model for 3d gaussian splatting.
\newblock In \emph{European Conference on Computer Vision}, pages 1--19. Springer, 2025.

\bibitem[Zhang et~al.(2024{\natexlab{b}})Zhang, Wang, Zhang, Qiu, Pang, Jiang, Yang, Xu, and Yu]{zhang2024clay}
Longwen Zhang, Ziyu Wang, Qixuan Zhang, Qiwei Qiu, Anqi Pang, Haoran Jiang, Wei Yang, Lan Xu, and Jingyi Yu.
\newblock Clay: A controllable large-scale generative model for creating high-quality 3d assets.
\newblock \emph{ACM Transactions on Graphics (TOG)}, 43\penalty0 (4):\penalty0 1--20, 2024{\natexlab{b}}.

\end{thebibliography}
}

\clearpage
\setcounter{page}{1}
\maketitlesupplementary

\section{Details of Methods }

\subsection{Physically-based Rendering Model}
\label{sec:rationale}

The traditional rendering equation~\cite{kajiya1986} is formulated following the principles of energy conservation in physics. It computes the outgoing radiance $L_o$ at surface point $\hat{\textbf{x}}$ with surface normal $\hat{\bm{n}}_x$ along view direction $\bm{\omega}_o$ by integrating over the hemisphere $\Omega^+=\{\bm{\omega}_i:\bm{\omega}_i\cdot\hat{\bm{n}}_x>0\}$, where $\bm{\omega}_i$ is the incident light direction:
\begin{equation}
    L_{o}(\hat{\textbf{x}}, \bm{\omega}_o) = \int_{\Omega^+}L_{i}(\hat{\textbf{x}},\bm{\omega}_i)f_r(\hat{\textbf{x}},\bm{\omega}_i,\bm{\omega}_o)(\bm{\omega}_i\cdot\hat{\bm{n}}_x)\text{d}\bm{\omega}_i \label{con:renderequation}
\end{equation}
\noindent The function $L_{i}(\hat{\textbf{x}},\bm{\omega}_i)$ represents the incoming radiance at the surface point $\hat{\textbf{x}}$ from $\bm{\omega}_i$, while the BRDF function $f_r$ quantifies the proportion of light arriving from direction $\bm{\omega}_i$ that is reflected in the direction $\bm{\omega}_o$ at point $\hat{\textbf{x}}$.

In Sec.3.2, We propose a \textbf{view-dependent illumination-aware condition} based on the microfacet BRDF model~\cite{karis2013real} to approximate surface reflectance properties using a set of decomposed intrinsic terms, such as Roughness and Metallic, which encapsulate the interaction of light with a surface. The microfacet model provides a robust framework for simulating the reflection from rough surfaces. The BRDF $f_r(\hat{\textbf{x}},\bm{\omega}_i,\bm{\omega}_o)$ is the ratio between the incoming and outgoing radiance. It is a function of the surface location $\hat{\textbf{x}}$, incoming direction $\bm{\omega}_i$, and the outgoing direction $\bm{\omega}_o$: 
\begin{equation}
    f_r(\hat{\textbf{x}},\bm{\omega}_i,\bm{\omega}_o) = f_d(\hat{\textbf{x}}) + f_s(\hat{\textbf{x}},\bm{\omega}_i,\bm{\omega}_o)
\end{equation}
\noindent where the BRDF comprises diffuse reflection $f_d$ and specular reflection $f_s$. In the first stage of BRDF estimation, we use the diffuse reflection term $f_d(\hat{\textbf{x}}) = \frac{\hat{A}_x}{\pi}$ based on a simple Lambertian model. Here, we introduce the specular BRDF $f_s$ which follows the Cook-Torrance model~\cite{cook1982reflectance}:
\begin{equation}
    f_s =  \frac{D(\bm{h}, \hat{\bm{n}}, R) \cdot F(\bm{\omega}_o, \bm{h}) \cdot G(\bm{\omega}_i, \bm{\omega}_o, \hat{\bm{n}}, R)}{4(\hat{\bm{n}} \cdot \bm{\omega}_o)(\hat{\bm{n}} \cdot \bm{\omega}_i)}
\end{equation}
\noindent where $\hat{\bm{n}}$ is the surface normal at $\hat{\textbf{x}}$, $R$ is roughness term, $\bm{h}$ is a half vector, $D$ denotes Normal Distribution Function (NDF), $F$ denotes Fresnel function and $G$ is the Geometry Factor.

The normal distribution function $D$ describes the probability density of microfacet orientations aligned with the half-vector $\bm{h}$. It is dependent on the macroscopic normal   $\bm{h}$ and the surface roughness $R$, influencing the distribution of specular highlights, as follows: 
\begin{equation}
    D(\bm{h},\hat{\bm{n}},R) = \frac{R^2}{\pi ((\hat{\bm{n}} \cdot \bm{h})^2 (R^2 - 1) + 1)^2}
\end{equation}
\noindent where the half-vector $\bm{h}$ is the normalized vector halfway between the incident light direction and the reflection direction, $\hat{\bm{n}}$ is the macroscopic surface normal, and $R=R^2$ is the surface roughness parameter, ranging from [0,1]. As $R$ approaches 0, the surface tends towards a mirror-like reflection, while $R$ approaching 1 indicates an extremely rough surface.

In the microfacet BRDF model, the Fresnel reflectance term \( F \) characterizes the proportion of light reflected from a material's surface at varying angles of incidence. This proportion changes with the angle of incidence, being lowest at normal incidence and highest at grazing angles for non-metallic materials. The Fresnel term can be efficiently computed using Schlick's approximation~\cite{schlick1994inexpensive}:
\begin{equation}
    F(\bm{\omega}_o, \bm{h}) = F_0 + (1 - F_0) \cdot (1 - (\bm{\omega}_o  \cdot \bm{h}))^5
\end{equation}
\noindent where $F_0$ is the reflectance at normal incidence, also known as the base reflectance and $\hat{\bm{\omega}_o} \cdot \bm{h}$ is the dot product between the outgoing radiance and the half-vector, dictating the cosine of the angle of incidence. $F_0$ is up to the intrinsic properties of the material, typically categorized as insulators or metals, and higher metallic leads to a higher $F_0$.

In the realm of Physically Based Rendering (PBR), particularly within the context of the GGX Specular reflection model, the Geometry Function \( G \) assumes a pivotal role. It is primarily responsible for simulating the occlusion and shadowing effects attributable to the microsurface structure, crucial for accurately rendering light reflections on rough surfaces. The geometry function is typically computed using a combination of the Smith geometry function and the Schlick-GGX approximation. This formula is written as:
\begin{equation}
    G(\hat{\bm{n}}, \bm{\omega}_o, \bm{\omega}_i, R) = G_1(\hat{\bm{n}}, \bm{\omega}_o, R) \cdot G_1(\hat{\bm{n}}, \bm{\omega}_i, R)
\end{equation}

\noindent Here, \( G_1 \) represents the monodirectional shadowing function:
\begin{equation}
    G_1(\hat{\bm{n}}, \bm{\omega}_o, R) = \frac{\hat{\bm{n}} \cdot \bm{\omega}_o}{(\hat{\bm{n}} \cdot \bm{\omega}_o)(1 - k) + k}
\end{equation}
\begin{equation}
    G_1(\hat{\bm{n}}, \bm{\omega}_i, R) = \frac{\hat{\bm{n}} \cdot \bm{\omega}_i}{(\hat{\bm{n}} \cdot \bm{\omega}_i)(1 - k) + k}
\end{equation}

\noindent where $R$ is the parameter indicating surface roughness, $k$ is a parameter derived from the roughness, typically calculated as  $k = \frac{R^2}{2}$.

\subsection{VLM-guided PBR Material generation}
For alleviating these hard ambiguities between Albedo and specular light derived from Albedo generation and due to Roughness and Metallic distribution for objects tend to be part-aware consistent, we leverage VLM to provide strong prior to alleviate these hard ambiguities. Specifically, we design the a two-round reflective-metalness material dialogue mechanism (RMD) by GPT-4V, which can effectively unleash the ability for reflective material knowledge. We observe that VLM not only accurately identifies whether materials are metallic or reflective but also determines material properties at the component level, as shown in the Fig.~\ref{fig:VLM_pipeline}. Using the CLIP~\cite{radford2021learning} model allows us to integrate this prior knowledge into the material generation process.

\section{Additional Results of Experiments}
\subsection{Mutil-view PBR Comparison results}
We show more results on text-to-mesh task compared with other baselines, see in Fig.~\ref{fig:pbr_compare1} and Fig.~\ref{fig:pbr_compare2}.

\begin{figure*}[t]
\centering
  \includegraphics[width=0.75\linewidth]{supp_figs/PBR_Comparison_1_supp.pdf}
  \vspace{-0.4cm}
\caption{Qualitative comparison with IntrinsicAnything~\cite{chen2024intrinsicanything} on multi-view PBR estimation. }
\label{fig:pbr_compare1}
\vspace{-0.4cm}
\end{figure*}

\begin{figure*}[t]
\centering
  \includegraphics[width=0.75\linewidth]{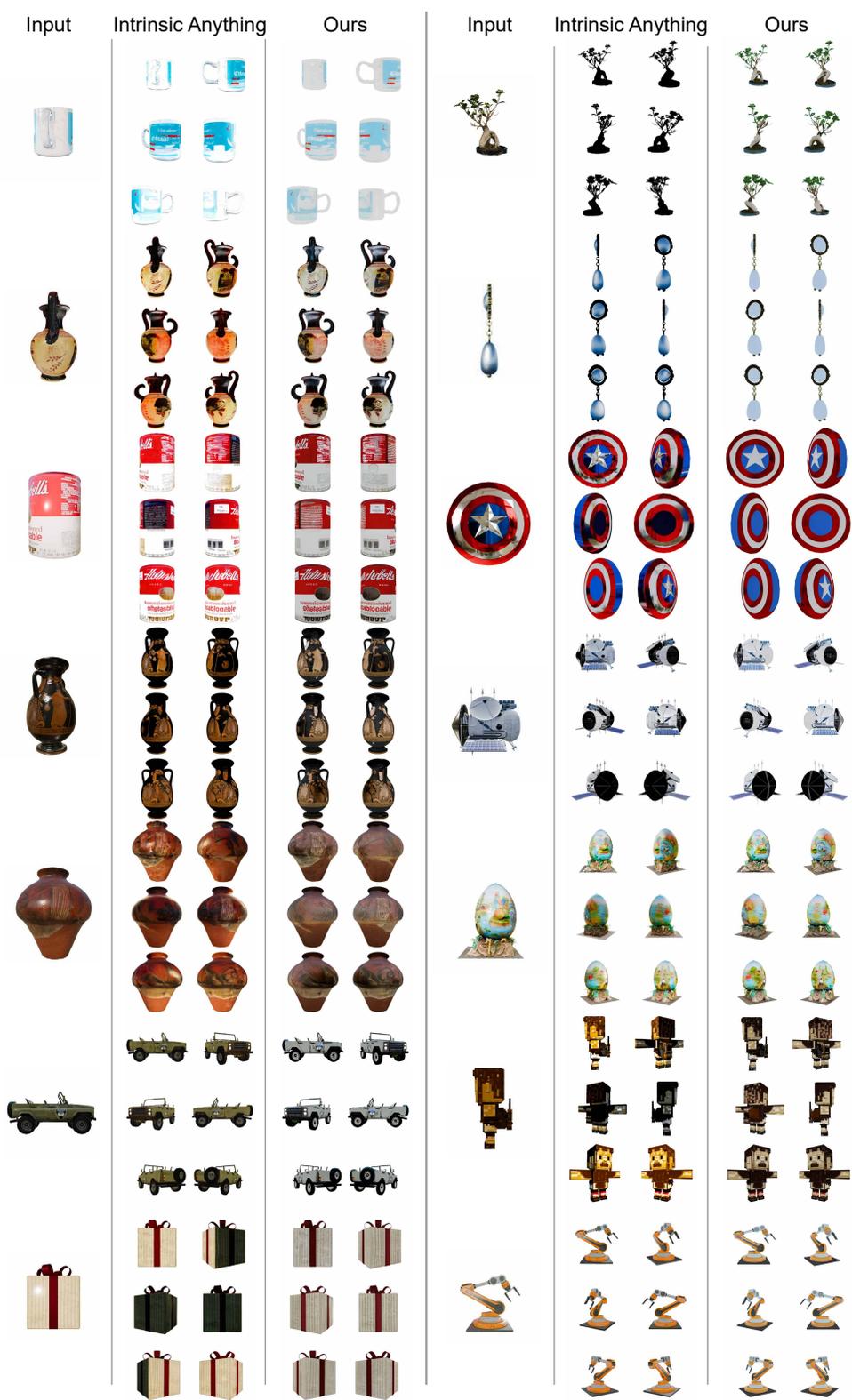}
  \vspace{-0.4cm}
\caption{Qualitative comparison with IntrinsicAnything~\cite{chen2024intrinsicanything} on multi-view PBR estimation. }
\label{fig:pbr_compare2}
\end{figure*}

\subsection{Text-to-mesh Comparison results}
As shown in Fig.~\ref{fig:text2mesh}, we also implement our method on text-to-mesh task compared with other baselines. The images are generated by HunyuanDiT\cite{huang2024dialoggen} conditioned on text prompts, and we remove their background.

\begin{figure*}[t]
\centering
  \includegraphics[width=0.88\linewidth]{supp_figs/comparison-3-supp.pdf}
  \vspace{-0.4cm}
\caption{Qualitative comparison of the generated 3D assets with other methods on Text-to-mesh.}
\label{fig:text2mesh}
\vspace{-0.4cm}
\end{figure*}

\subsection{Image-to-mesh Comparison results}
We show more results on text-to-mesh task compared with other baselines, see in Fig.~\ref{fig:image2mesh1} and Fig.~\ref{fig:image2mesh2}.

\begin{figure*}[t]
\centering
  \includegraphics[width=0.88\linewidth]{supp_figs/comparison-2-supp.pdf}
  \vspace{-0.4cm}
\caption{Qualitative comparison of the generated 3D assets with other methods on Image-to-mesh. }
\label{fig:image2mesh1}
\end{figure*}

\begin{figure*}[t]
\centering
  \includegraphics[width=0.88\linewidth]{supp_figs/comparison-1-supp.pdf}
  \vspace{-0.4cm}
\caption{Qualitative comparison of the generated 3D assets with other methods on Image-to-mesh.}
\label{fig:image2mesh2}
\vspace{-0.4cm}
\end{figure*}

\subsection{Abaltion results for Multi-view PBR estimation}
We show more results on abaltion study results for Albedo and MRO maps, see in Fig.~\ref{fig:ablation_albedo} and Fig.~\ref{fig:ablation_mro}.

\subsection{Comparison results for PBR estimation}
Given the lack of large-scale PBR datasets and the difficulty in obtaining real-world PBR, we used the only publicly available PBR dataset from Objaverse for intrinsic image evaluation. Additionally, we evaluated our RGB and geometry metric on the GSO dataset, outperforming existing methods. To further validate our results, we add FID metrics on both Objaverse and GSO datasets in Table.~\ref{re_table:FID}.

\subsection{Robust experiments for PBR estimation:}
\begin{figure*}[htbp]
\centering
  \includegraphics[width=1\linewidth]{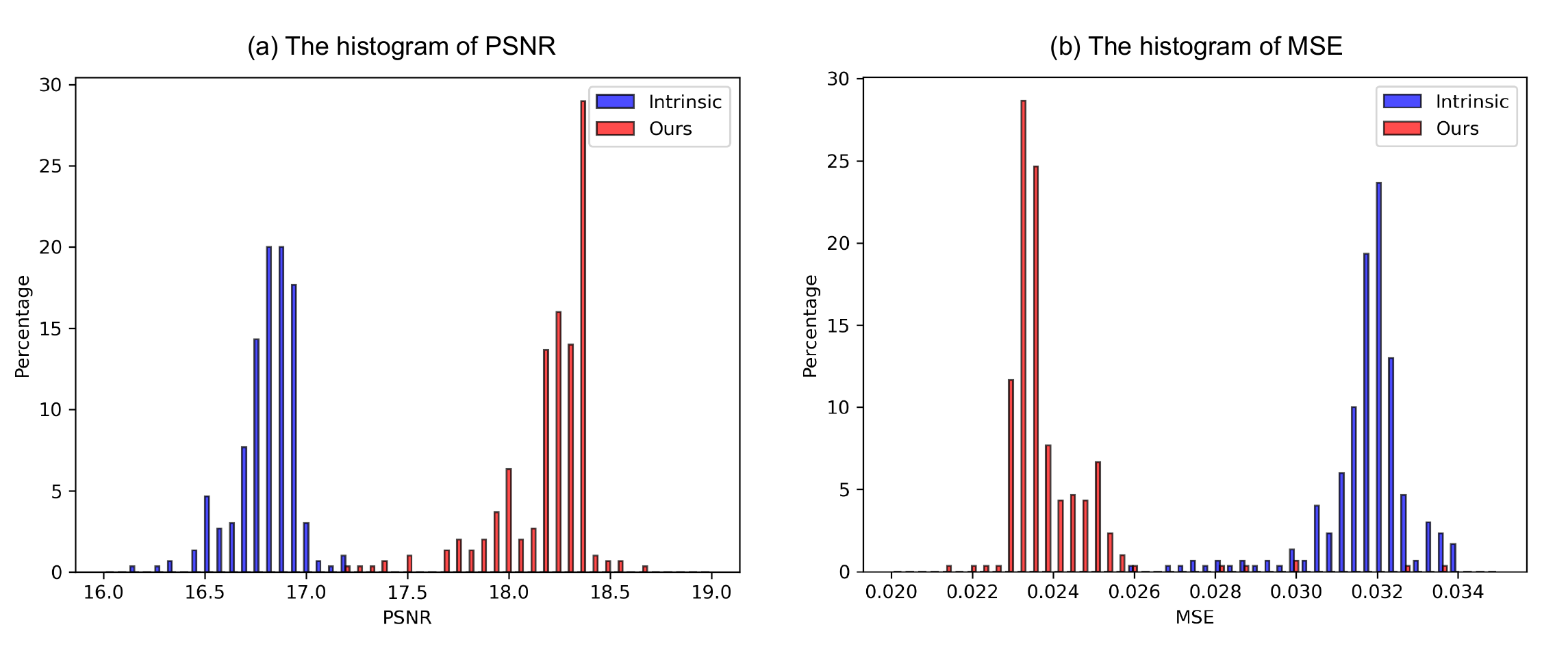}
  \vspace{-0.4cm}
\caption{ Quantitative results on PBR estimation robust experiments.}
\label{re_fig:robustness}
\end{figure*}

We add the distribution of predicted albedo in Fig.~\ref{re_fig:robustness}(a). 
Our outperforms IntrinsicAnything with 8.4\% higher PSNR and 25.5\% lower MSE, alongside \textbf{5.7\%} and \textbf{30.9\%} reductions in \textbf{Standard Deviation}, respectively, demonstrating superior robustness and accuracy.

\begin{figure*}[htbp]
\vspace{-0.5cm}
\centering
  \includegraphics[width=0.850\linewidth]{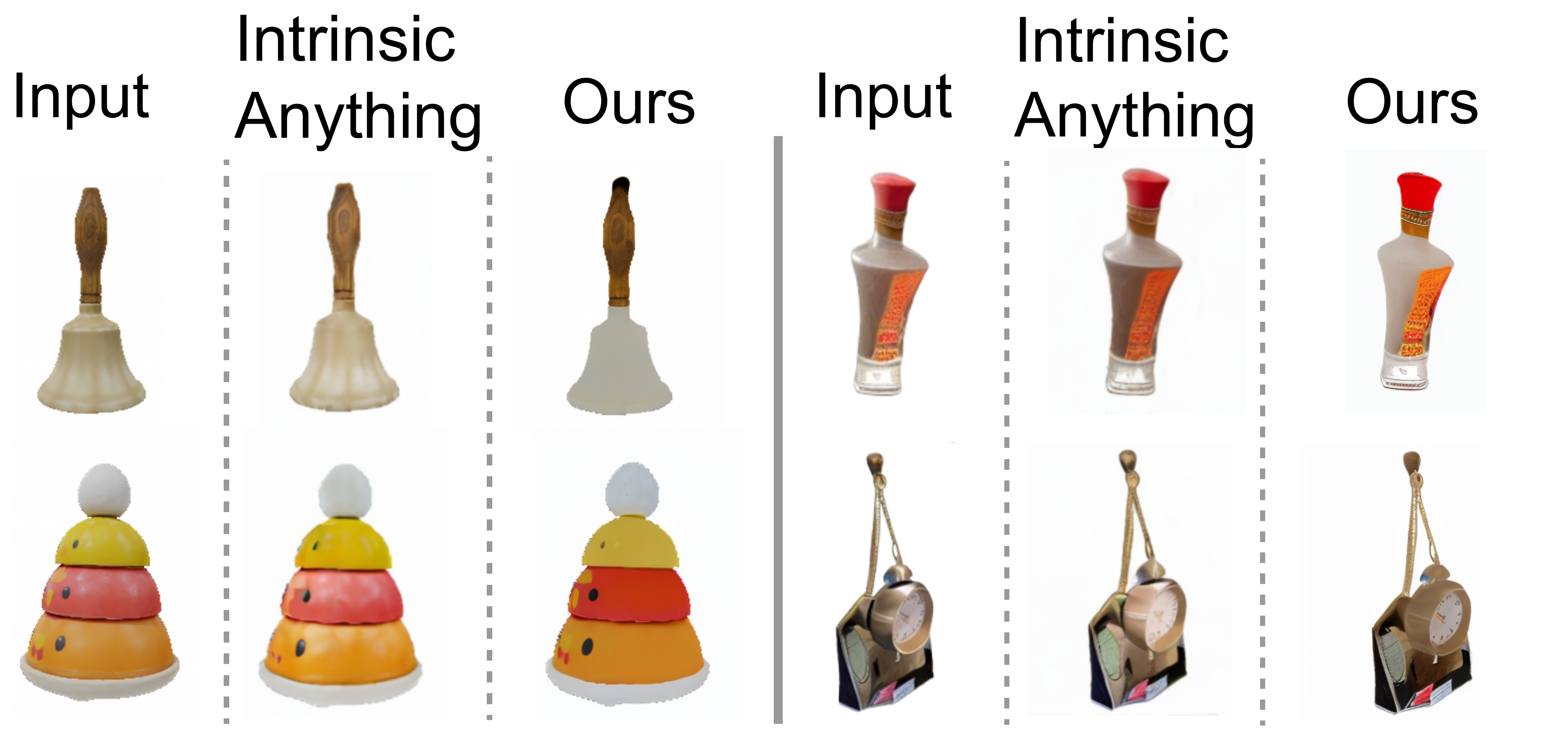}
  \vspace{-0.4cm}
\caption{ Qualitative PBR results on real-world data.}
\label{re_fig:in_the_wild_compare}
\end{figure*}

Meanwhile, we test our PBR diffsuion model on two real-world dataset, GSO and MVImgNet. Our method outperforms IntrinsicAnything, especially on reflective surfaces, reducing lighting and shadow ambiguity in Fig.~\ref{re_fig:in_the_wild_compare}.\\

\subsection{Method of feature injection:}
Our injects features via latent noise concatenation, outperforming IP-Adapter and ControlNet in pixel-level alignment of image contours and structures, as shown in Table.~\ref{re_table:inject_compare}. 

\begin{table}[htbp]
\centering
\caption{Comparison of different injection methods.}
\vspace{-0.35cm}
\resizebox{1\linewidth}{!}{
\begin{tabular}{@{}lllllll@{}}
\toprule
\multirow{2}{*}{Methods} & \multicolumn{2}{c}{Albedo} & \multicolumn{2}{c}{Metallic} & \multicolumn{2}{c}{Roughness} \\
\cmidrule(lr){2-3} \cmidrule(lr){4-5} \cmidrule(l){6-7}
& PSNR & MSE & PSNR & MSE & PSNR & MSE \\
\midrule
ControlNet & 17.874 & 0.025 & 16.451 & 0.039 & 16.960 & 0.031 \\
Ip-adapter & 17.351 & 0.029 & 16.546 & 0.038 & 15.932 & 0.036 \\
\textbf{Concat(Ours)} & \textbf{18.186} & \textbf{0.023} & \textbf{17.718} & \textbf{0.028} & \textbf{21.095} & \textbf{0.013} \\
Five-channel & 16.986 & 0.033 & 17.118 & 0.031 & 20.985 & 0.016 \\
\bottomrule
\end{tabular}
}
\label{re_table:inject_compare}
\end{table}

In PBR rendering, metallic/roughness control reflective effects while albedo defines base color. These material properties are independent: an object may share metallic values but differ in color. Using separate latent spaces aligns with physical rendering principles. Additionally, specular light condition affects only metallic/roughness, and independent encoding avoids interference with albedo sampling. The results in Table.~\ref{re_table:inject_compare} further validate our idea. 

\subsection{Specular illumination maps:}
\begin{figure*}[htbp]
\centering
  \includegraphics[width=1\linewidth]{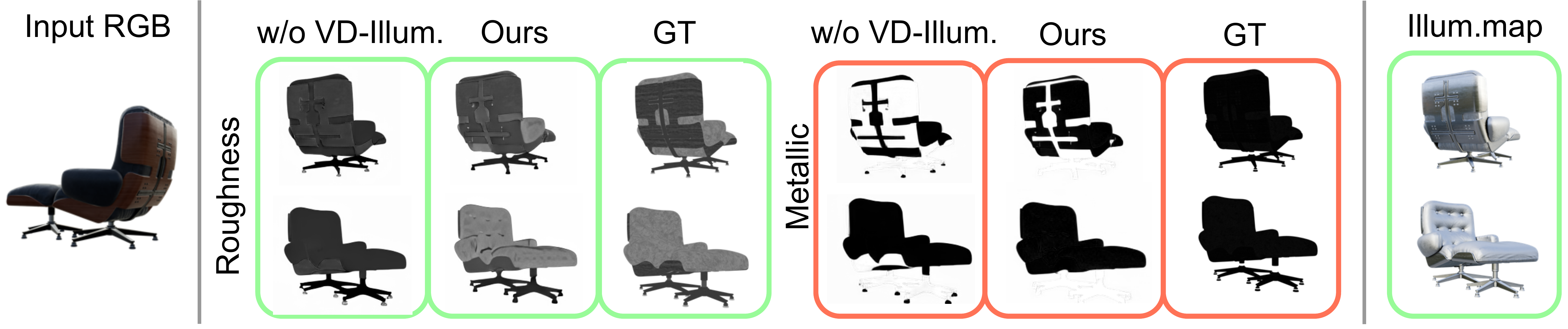}
  \vspace{-0.35cm}
\caption{ The qualitative result for specular light condition.}
\label{re_fig:spec_light}
\end{figure*}

\begin{figure*}[t]
\centering
  \includegraphics[width=0.95\linewidth]{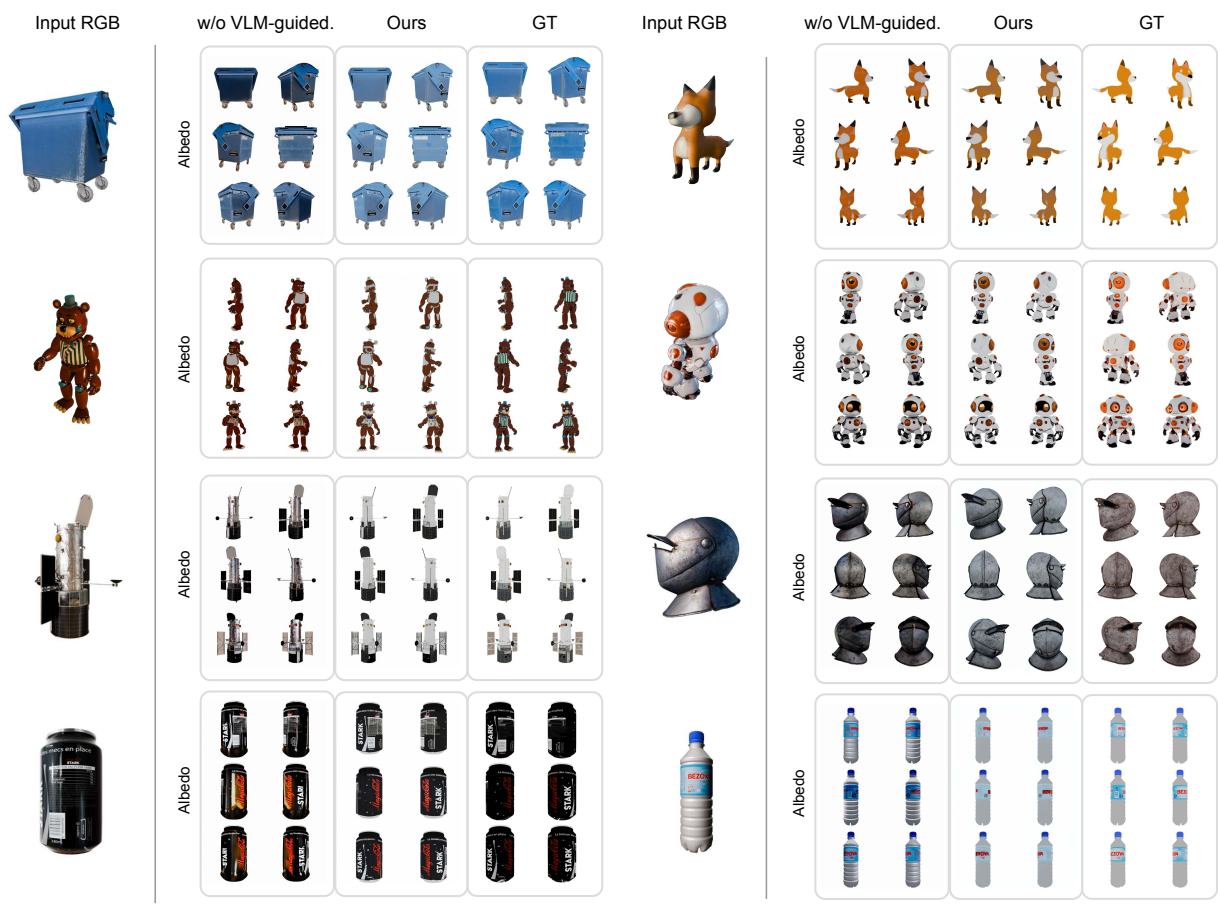}
  \vspace{-0.4cm}
\caption{Ablation studies for Mulit-view albedo estimation. All scores are calculated
as an average across 300 objects from the Objaverse dataset. }
\label{fig:ablation_albedo}
\end{figure*}

We directly rendered the specular light map using Blender as GT. As shown in Fig.~\ref{re_fig:spec_light}, the distinct highlights on the chair back and leather regions in the specular light map guide spatially varying roughness generation and further improve roughness/metallic accuracy. \\

\vspace{-0.4cm}
\begin{figure*}[t]
\centering
  \includegraphics[width=0.75\linewidth]{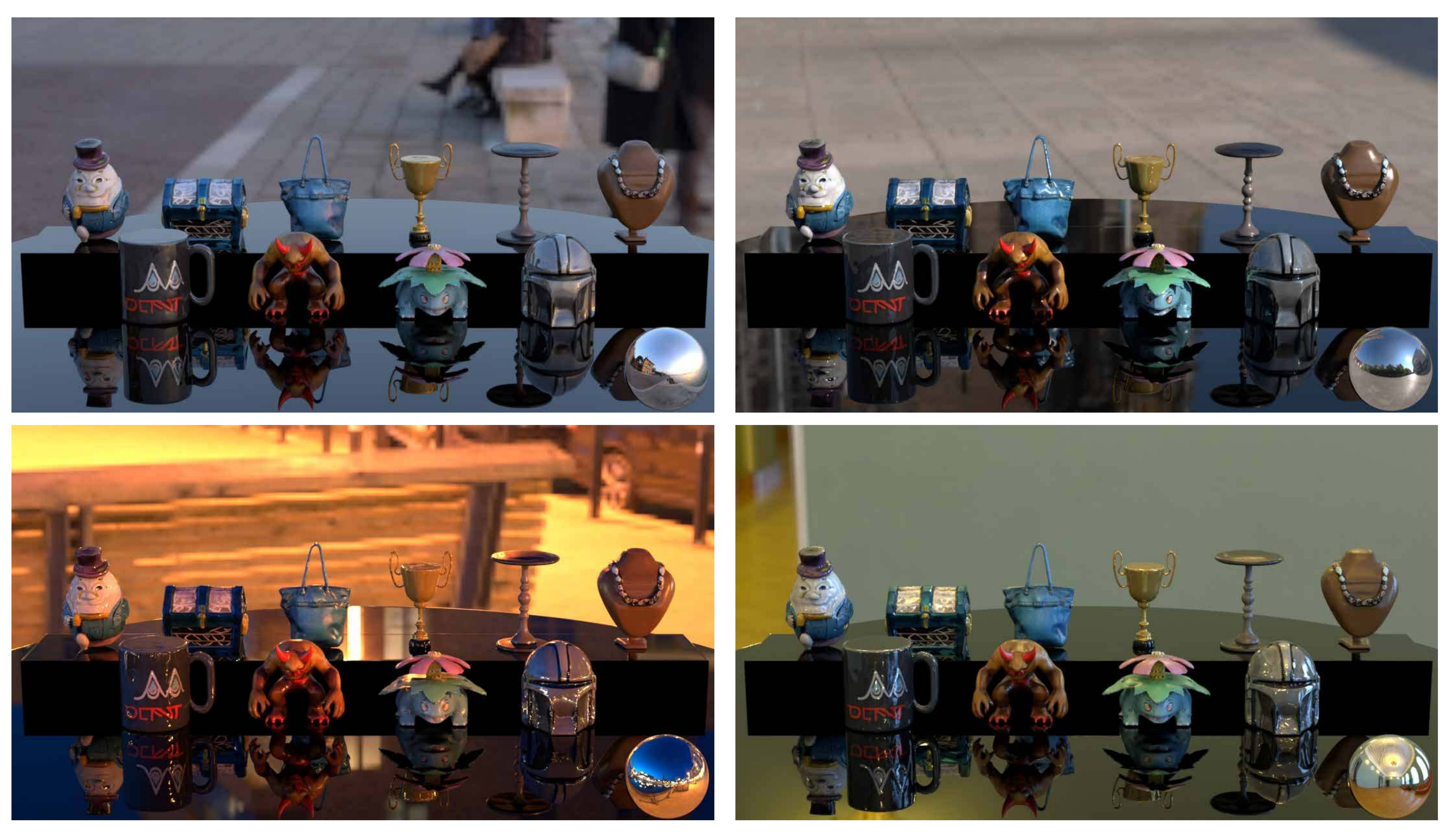}
  \vspace{-0.0cm}
\caption{Relighting results under different environment illumination. }
\label{fig:relighting}
\vspace{-0.4cm}
\end{figure*}

\begin{figure*}[t]
\centering
  \includegraphics[width=0.85\linewidth]{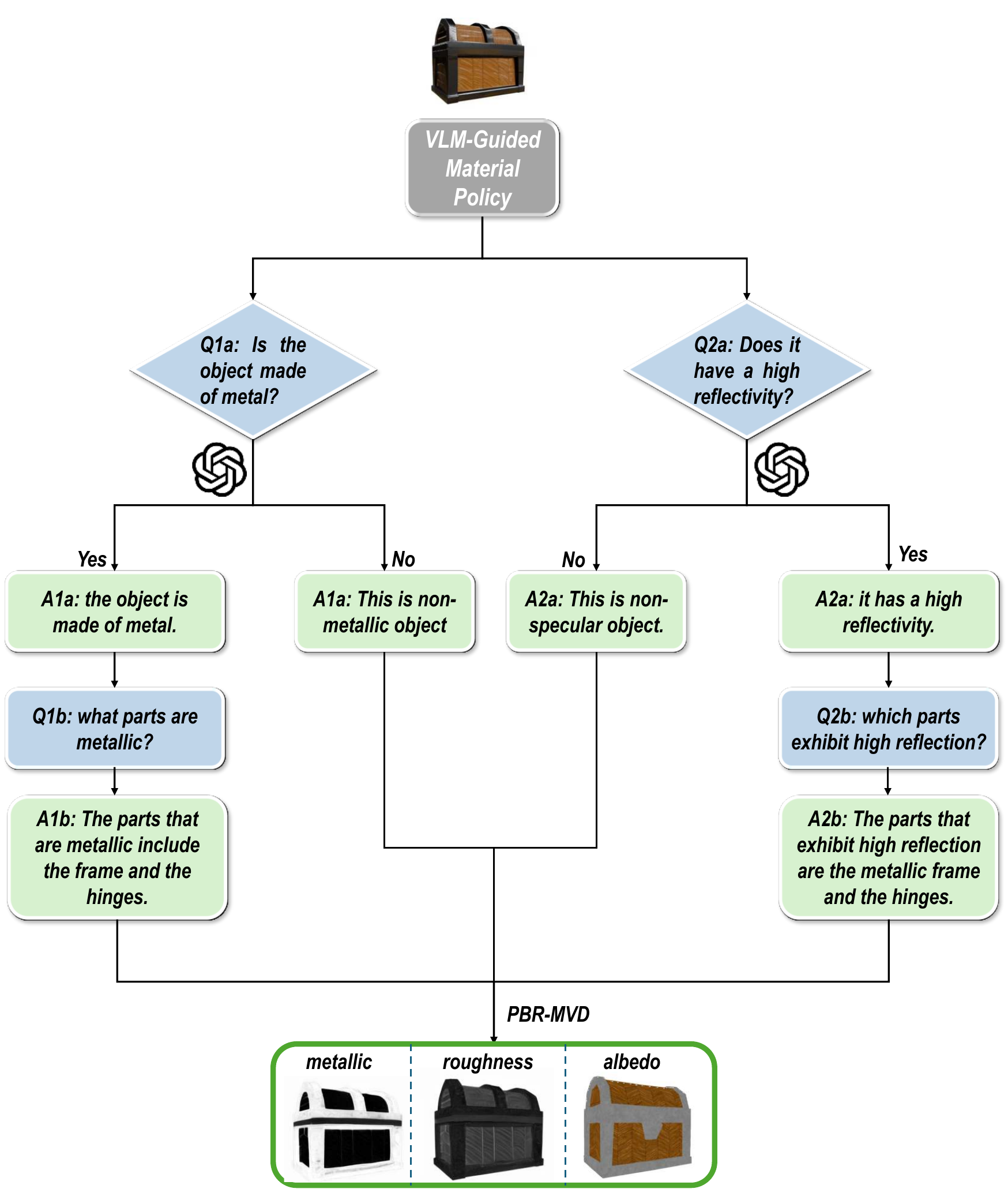}
  \vspace{-0.0cm}
\caption{The VLM-Guided material policy, including two-round reflective-metalness material dialogue mechanism. }
\label{fig:VLM_pipeline}
\vspace{0.4cm}
\end{figure*}

\begin{figure*}[t]
\centering
  \includegraphics[width=0.6\linewidth]{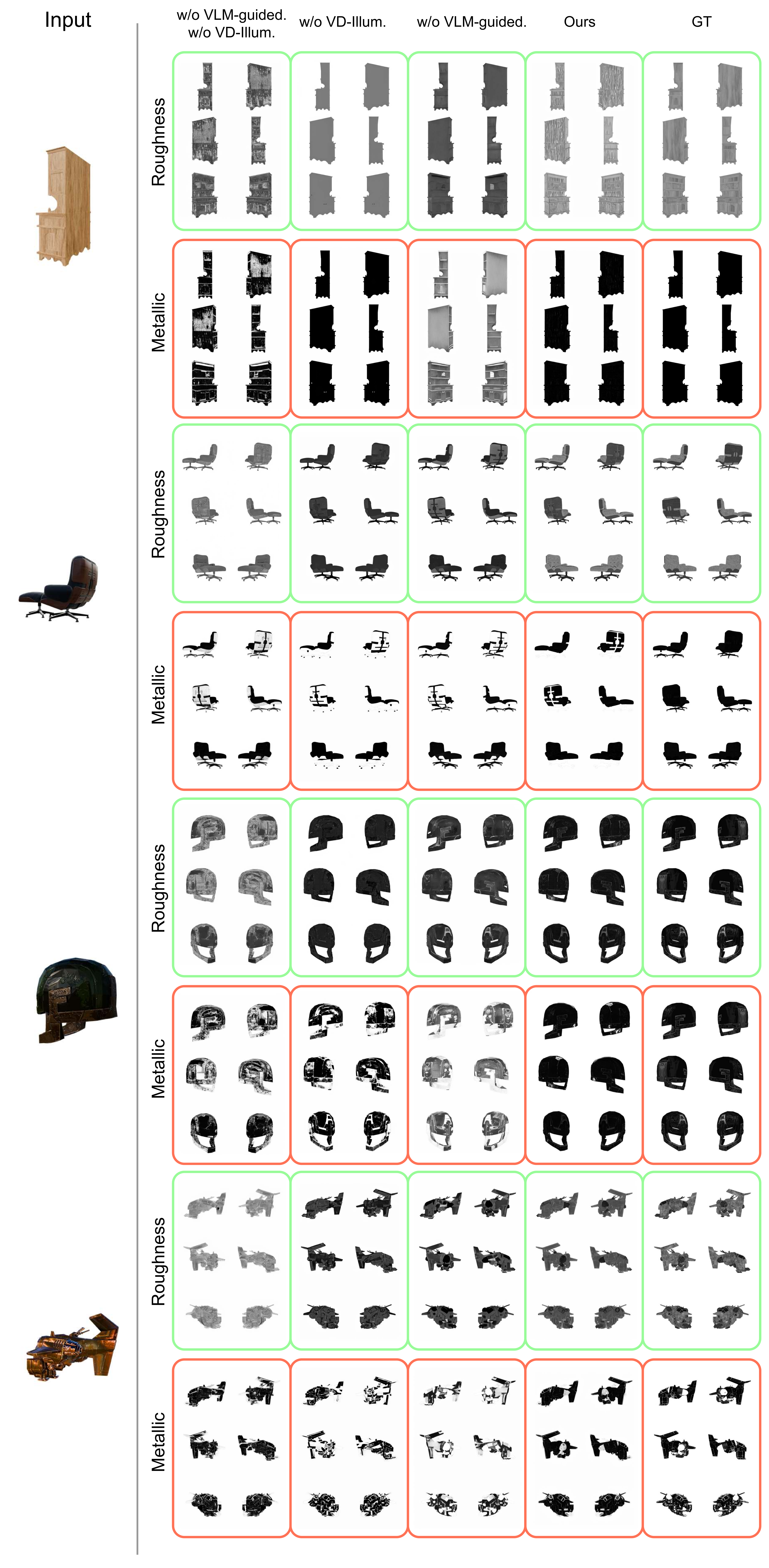}
  \vspace{-0.4cm}
\caption{Ablation studies for Mulit-view metallic and roughness estimation. All scores are calculated
as an average across 300 objects from the Objaverse dataset. }
\label{fig:ablation_mro}
\end{figure*}

\section{Applications}
Based on our PBR model, we can further relight and animate the model under different illumination, as shown in Fig.~\ref{fig:relighting}. More video results and visualization can be found on our project page: 
\href{https://pbr3dgen1218.github.io/}{https://pbr3dgen1218.github.io/}.

\begin{table}[htbp]
\centering
\caption{FID($\downarrow$) comparison on GSO and Objaverse.}
\vspace{-0.35cm}
\resizebox{1\linewidth}{!}{
\begin{tabular}{@{}lccccccc@{}}
\toprule
Methods & Ours & SF3D & PrimX & OpenLRM & TripoSR & InstantMesh & LGM \\ 
\midrule
RGB(GSO) & \textbf{65.04} & 87.32 & 75.86 & 113.42 & 109.32 & 91.32 & 132.64 \\
RGB(Objaverse) & \textbf{49.37} & 55.36 & 68.49 & - & - & - & - \\
Albedo(Objaverse) & \textbf{50.63} & 57.67 & 78.55 & - & - & - & - \\
\bottomrule
\vspace{-1.07cm}
\end{tabular}
}
\label{re_table:FID}
\end{table}



\end{document}